%% file: main.tex
\documentclass[times,review,preprint]{elsarticle}

\usepackage{framed,multirow}
\usepackage{url}
\usepackage{xcolor}

\usepackage{amsmath}
\usepackage{amssymb}
\usepackage{amsthm}
\usepackage{latexsym}
\usepackage{lineno}
\usepackage[colorlinks]{hyperref}
\usepackage[utf8]{inputenc}
\usepackage[english]{babel}
\usepackage[font=small,labelfont=bf]{caption}
\usepackage[font=small,labelfont=bf]{subcaption}
\usepackage{graphicx}
\usepackage{setspace}
\usepackage[ruled]{algorithm2e}

{}
\newtheorem{corollary}{Corollary}{}
{}
\newtheorem{proposition}{Proposition}{}
\newtheorem{definition}{Definition}{}

\newcommand{\scorefunction}{\mathcal{F}_{\Omega}}

\begin{document}


\begin{frontmatter}

\title{Cosine-Pruned Medial Axis: A new method for isometric equivariant and noise-free medial axis extraction}


\author[1,2]{Diego Patiño}\corref{cor1}
\author[2]{John Branch}

\cortext[cor1]{Corresponding author: e-mail: \href{diegopc@seas.upenn.edu}{diegopc@seas.upenn.edu}}

\address[1]{School of Engineering and Applied Science, University of Pennsylvania, 220 South 33rd St., Philadelphia, PA, 19104, USA}
\address[2]{Faculty of Mines, National University of Colombia, Av. 80 \# 65 - 223, Medellín, Colombia}

\begin{abstract}

We present the CPMA, a new method for medial axis pruning with noise robustness and equivariance to isometric transformations. Our method leverages the discrete cosine transform to create smooth versions of a shape $\Omega$. We use the smooth shapes to compute a score function $\scorefunction$ that filters out spurious branches from the medial axis. We extensively compare the CPMA with state-of-the-art pruning methods and highlight our method's noise robustness and isometric equivariance. We found that our pruning approach achieves competitive results and yields stable medial axes even in scenarios with significant contour perturbations.

\end{abstract}

\begin{keyword}
Medial Axis Pruning \sep Discrete Cosine Transform \sep Equivariance \sep Isometric transformation \sep Morphological skeleton
\end{keyword}

\end{frontmatter}



\section{Introduction}
\input{sections/01_introduction}

\section{Related work}
\input{sections/02_related_work}

\section{Method}
\input{sections/03_proposed_method}

\section{Experiments}
\input{sections/04_experiments}

\section{Results}
\input{sections/05_results}

\section{Conclusions and Future Work}
\input{sections/06_conclusions}

\section{Acknowledgments}

We want to thank the GRASP Lab at the University of Pennsylvania for providing the computational resources to conduct this investigation.

\begin{scriptsize}
\setcitestyle{square}
\bibliographystyle{plain}
\bibliography{bib}
\end{scriptsize}

\end{document}

%% file: sections/01_introduction.tex
Shape analysis arises naturally in computer vision applications where geometric information plays an essential role. The shape of an object is a useful tool in fields such as: non-destructive reconstruction of archaeology and cultural heritage~\citep{Tal2014,Maaten2006}; object classification and retrieval from large collections of images~\citep{Li2018,Safar2003}; human action and pose recognition for gaming and entertainment~\citep{Chaudhry2013,Li2019ActionalStructuralGC}; environment sensing in robot navigation and planning~\citep{Peters2016,Li2017abc}; and industry for automatic visual quality inspection of product defects~\citep{6005551,7071593}. 

We visually perceive shape as the collections of all the features that constitute an object. However, to perform computer-based shape analysis, one must rely on an accurate discrete mathematical representation of an object's shape. This representation should exhibit the same geometric and topological properties inherent to the shape itself. Accordingly, we can think of shape representation as a way to store the shape's information in a different format, which benefits speed, compactness, and efficiency.

Many authors have proposed a variety of shape representations such as voxel/pixel grids, point clouds, triangular meshes, medial axes, or signed distance functions~\cite{Siddiqi1999,Toshev2012,Marie2016,Black2016,Zhang2004ab}. These representations differ greatly in their formulation, and aim to provide a method for extracting descriptive features from objects, while also preserving their appearance and geometric properties~\cite{Belongie2002,Abbasi1999,Sun2009,Esteves2018a,Maturana2015}. However, these methods also have disadvantages that limit their application. For example, medial axis representations are highly sensitive to contour noise; voxel/pixel grids are inaccurate after isometric transformations; signed distance functions and triangular meshes are memory-consuming representations when high-frequency details of the shape want to be stored.

We focus this study on the medial axis, also called the topological skeleton. The medial axis represents shapes as a collection of one-dimensional curves that define the central axis (or backbone) of an object. It provides dimensionality reduction of the amount of data needed to represent an entire shape while preserving its topological structure. Moreover, the medial axis is a rotation equivariant shape representation because the medial axis of a rotated object should ideally be the rotated medial axis of the same object. The medial axis is also robust to small deformation, such as articulation, because of its graph-like structure. For instance, a human-like shape moving only its arm will not affect all of the points in the medial axis, only the connections between the arm's nodes.

Despite its advantages, the medial representation is extremely sensitive to noise on the object's contour~\cite{Saha2016,Saha2017}. Even small amounts of noise can cause erroneous sections of the skeleton called \textit{spurious branches}. Consequently, many medial axis extraction algorithms are equipped with a mechanism to avoid or remove these spurious branches. There are two main strategies reported in the state-of-the-art to deal with this problem: prior smoothing of the curve representing the object's boundary, and pruning the spurious branches after the medial axis' computation. In the former, the smoothed boundary is obtained by removing small structures along the curve or surface. It is interesting to note that smoothing curves does not always result in a simplified skeleton~\cite{younes2019,August99}. Effective pruning techniques focus instead on criteria to evaluate the significance of individual medial axis branches. However, pruning often requires user-defined parameters that depend on the size and complexity of the object~\cite{SHAKED1998156,Bai2007ab,Shen2011AB}, making the pruning method domain-dependent. Moreover, some pruning strategies result in a violation of the equivariant property~\cite{Saha2017,Saha2016}. As a result, medial axis pruning is still an open problem in computer vision, and this problem is in need of noise-robust methods that concurrently preserve the isometric equivariance of the medial axis. 

This paper presents a new method for medial axis pruning that employs mechanisms from the two aforementioned branch-removal strategies. Our method works by computing a controlled set of smoothed versions of the original shape via the discrete cosine transform. We combine these smoothed shapes' medial axis to create a score function that rates points and branches by their degree of importance. We use our score function to prune spurious branches while preserving the medial axis' ability to reconstruct the original object. Our method is robust to boundary noise and exhibits isometric equivariance.

We benchmark our approach on three datasets of 2D and 3D segmented objects. We use the Kimia216~\citep{Sebastian2004} and the Animal dataset~\citep{Bai2009} to evaluate 2D medial axis extraction. These two datasets provide a method to assess 2D medial axis extraction in the presence of intra-class variation. We also use the University of Groningen Benchmark~\citep{Sobiecki2014,Sobiecki2013,Chaussard2011} to evaluate our approach on 3D objects. Our results show that our approach achieves competitive results on isometric equivariance and noise robustness compared to the state-of-the-art.

The main \textbf{contributions} of this paper are summarized as follows:

\begin{itemize}
    \item We define a novel method to compute medial axes that are robust to several degrees of boundary noise without losing the capacity to reconstruct the original object.
    \item Our computation pipeline guarantees that the isometric equivariance of the medial axis is preserved.
    \item The definition of our score function allows for a medial axis pruning that is efficiently computed in parallel.
\end{itemize}

%% file: sections/02_related_work.tex
Many algorithms and strategies exist to extract the medial axis and simplify it when affected by contour noise. This section briefly reviews the most representative algorithms for medial axis computation and discusses their key advantages and disadvantages.

\subsection{The Medial Axis}

Blum~\cite{Blum1967} first introduced the medial axis as an analogy of a fire propagating with uniform velocity on a grass field. The field is assumed to have the form of a given shape. If one ``sets fire'' at all boundary points, the medial axis is the set of quench points. There are other equivalent definitions of the medial axis. In this work we use a geometric definition as follows:

\begin{definition}{\textbf{Medial Axis.}}\label{def:MAT} Let $\Omega$ be a connected bounded domain in $\mathbb{R}^{n}$, and $x, x^{\prime}$ two points such that $x, x^{\prime} \in \Omega$. The medial axis of $\Omega$ is defined as all the points $x$ where $x$ is the center of a maximal ball $B_{r}$ of radius $r$ that is inscribed inside $\Omega$. Formally,

\begin{displaymath}
     \mathbf{MA}(\Omega) = \left\{ x \mid B_{r}(x) \not \nsubseteq B_{r ^{\prime}}(x^{\prime}), \forall r^{\prime} > r  \right \}.
\end{displaymath}
\end{definition}

\begin{figure}
    \centering
        \includegraphics[width=0.7\linewidth]{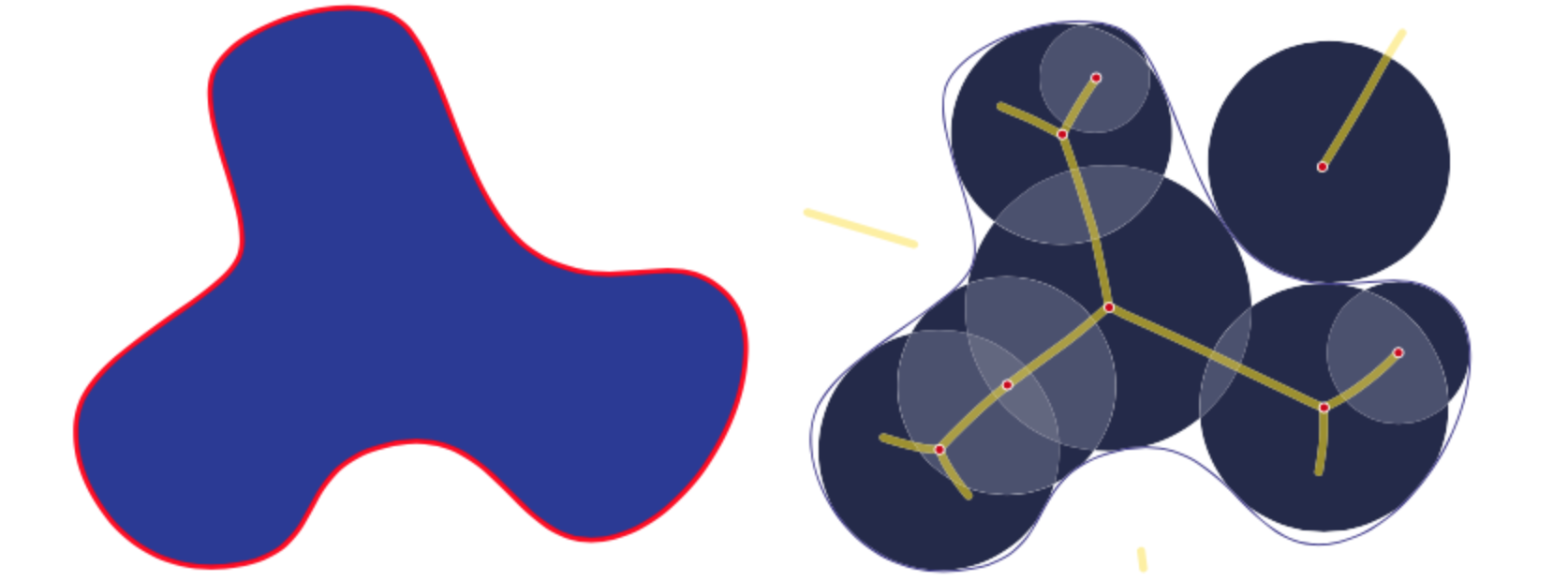}
        \caption[Medial Axis Transform computation]{Medial Axis Transform Computation. (Left) a shape and its boundary. (Right) Medial Axis elements consisting of the centers and radius of balls inscribed in the shape~\citep{Peters2016}.}
    \label{fig:medial_axis_visual_example}
\end{figure}

The medial axis, together with the associated radius of the maximally inscribed ball, is called the medial axis transform ($\mathbf{MAT}(\Omega)$). The medial axis transform is a complete shape descriptor, meaning that it can be used to reconstruct the shape of the original domain. In some work, $\mathbf{MA}$ and $\mathbf{MAT}$ are also referred to as shape skeletonization. Figure \ref{fig:medial_axis_visual_example} shows an example of a 2D shape and its medial axis as the center of maximal discs. In $\mathbb{R}^{3}$ definition \ref{def:MAT} may result in a 2-dimensional medial axis sometimes called the medial surface. We will restrict our examples and analysis to only one-dimensional medial axes.

\subsection{Medial Axis Computation}

There are three primary mechanisms to compute the $\mathbf{MA}$: 1) layer by layer morphological erosion also called thinning methods, 2) extraction of the medial axis from the edges of the Voronoi diagram generated by the boundary points, and 3) detection of ridges in distance map of the boundary points. In digital spaces, only an approximation to the ``true medial axis'' can be extracted. 

When using thinning methods ~\cite{Dlotko14,LOHOU2004171,5396343,Nemeth2011}, points which belong to $\Omega$ are deleted from the outer boundary first. Later, the deletion proceeds iteratively inside until it results in a single-pixel wide medial axis. The medial axis by thinning can be approximated in terms of erosion and opening morphological operations~\citep{Zhang1984}. Thinning methods are easy to implement in a discrete setting, but they are not robust to isometric transformations. 



The most well-known algorithm for thinning skeletonization is perhaps the Zhang Suen~\citep{Zhang1984} algorithm. However, other approaches have been developed using similar principles~\citep{Viswanathan2013,5396343,Nemeth2011}, mainly focused on parallel computation.

Another way to estimate the medial axis works by computing the Voronoi Diagram (VD) of a polygonal approximation of the object's contour. The contour is expressed as line segments in 2D or a polygonal mesh in 3D. The seed points for the VD are the vertices of the polygonal representation. The medial axis is then computed as the union of all of the edges $e_{ij}$ of the VD, such that the points $i$, and $j$ are not neighbors in the polygonal approximation~\citep{Ogniewicz1992}. 

Voronoi skeletonization methods preserve the topology of $\Omega$. However, a suitable polygonal approximation of an object is crucial to guarantee the medial axis' accuracy. Noise in the boundary forms convex areas in the contour, which induce spurious branches on the medial axis. In general, the better the polygonal approximation, the closer the Voronoi skeleton will be to the real medial axis. Nevertheless, this is an expensive process, particularly for large and complex objects~\citep{Saha2017}.

The most common methods to extract the medial axis are those based on the distance transform. Within these methods, the medial axis is computed as the ridges of the distance transform inside the object~\cite{Arcelli2011,Saha2016,ARCELLI1988361,Hesselink2008,Couprie2007,Postolski2014,Sato2000}. This interpretation of the medial axis follows definition \ref{def:MAT}, because the centers of the maximal balls are located on points $x$ along the ridgeline of the distance transform, and the radius of the balls correspond to the distance value at $x$. 

When computed in a discrete framework, distance-based approaches suffer from the same isometric robustness limitations as thinning and Voronoi methods ~\citep{Saha2017}. Moreover, noise in the contour produced by a low discretization resolution directly affects the medial axis' computation by introducing artificial ridges in the distance transform and, consequently, spurious branches.

\begin{figure}
    \centering
    \includegraphics[width=0.7\linewidth]{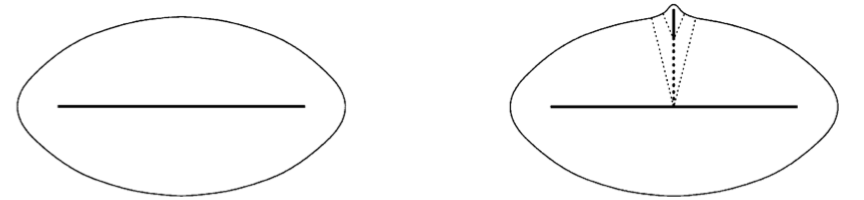}
    \caption[Spurious branch in medial axis.]{(Left) Spurious branch in medial axis. (Right) A new branch appears in the presence of a small perturbation in the contour~\cite{SHAKED1998156}.}
    \label{fig:example_spurious_branch}
\end{figure}

\subsection{Medial Axis Simplification}

The medial axis' sensitivity to boundary noise limits its applications to real-life problems~\citep{5395557}. Even negligible boundary noise can cause spurious branches, as shown in Figure \ref{fig:example_spurious_branch}. 

One strategy for removal of spurious branches consists of computing $\mathbf{MA}(\Omega)$ as the medial axis of a smoother version of the shape, $\Omega^{\prime}$~\citep{Rumpf2002,Mokhtarian1992789}. The main disadvantage of this approach is that, in most cases, the resulting medial axis is not a good approximation. Additionally, the smoothing of $\Omega$ can potentially change its topology.  Miklos et al.~\cite{Miklos:2010:DSA:1778765.1778838} introduced a slightly different approach they call Scaled Axis Transform (SAT). The SAT involves scaling the distance transform and computing the medial axis of the original un-scaled shape as the medial axis of the scaled one. However, in \citep{Postolski2014}, the authors show that the SAT is not necessarily a subset of the medial axis of the original shape. Instead, they propose a solution that guarantees a better approximation by including additional constraints on the scaled distance transform.

Another method to overcome the noise sensitivity limitation of the medial axis is spurious branch pruning. Pruning methods are a family of regularization processes incorporated in most medial axis extraction algorithms~\cite{SHAKED1998156}. Effective pruning techniques focus on different criteria to evaluate the significance of medial axis branches. Thus, the decision is whether to remove the branch (and its points) or not. We can say that pruning methods are adequate if the resulting $\mathbf{MA}$ is noticeably simplified while preserving the topology and its ability to reconstruct the original object. Most pruning methods rely on \textit{ad hoc} heuristic rules, which are invented and often reinvented in a variety of equivalent application-driven formulations~\cite{SHAKED1998156}. Some authors apply these rules while computing all medial axis points. Others do so by removing branches that are considered useless after the computation~\citep{Saha2017,Gao2018,Hesselink2008}.

One of the most popular pruning methods is Couprie et al.~\citep{Couprie2007}. They consider the angle $\theta$ formed by a point $x \in \Omega$, and its two closest boundary points denoted by the set $\Pi_{\Omega}(x)$. This solution removes points from $\mathbf{MA}$ for which $\theta$ is lower than a constant threshold. This criterion allows different scales within a shape but generally leads to an unconnected medial axis. Another pruning method found in the state-of-the-art is the work of Hesselink et al.~\citep{Hesselink2008}. They introduce the Gamma Integer Medial Axis (GIMA), where a point belongs to the medial axis if the distance between its two closest boundary points is at least equal to $\gamma$.

Many pruning methods rely on the distance transform $\mathcal{D}_{\Omega}(x)$. The distance transform acts as a generator function for the medial axis, such that points $x \in \mathbf{MA}$ if and only if they satisfy some constraint involving their distance to the boundary. However, other authors have proposed alternative generator functions in their pruning strategies~\cite{Gorelick2006,Aubert2014}. 

For example, \citep{Gorelick2006} and \citep{Aubert2014} introduce what they call Poisson skeletons by approximating $\mathcal{D}_{\Omega}(x)$ as the solution of the Poisson equation with constant source function. Poisson skeletons rely on a solid mathematical formulation. Among other concepts, they use the local minimums and maximums of the curvature of $\delta \Omega$. However, when such methodology is applied in a discrete environment, many spurious branches appear due to the need to define the length of a kernel size to estimate these local extreme points. 


%% file: sections/03_proposed_method.tex
\label{sec:proposed_method}

We propose a new pruning approach to remove spurious branches from the medial axis of a $n$-dimensional closed shape $\Omega$. We call our method the Cosine-Pruned Medial Axis (CPMA). The CPMA works by filtering out points from the Euclidean Medial Axis $\mathbf{MAT}(\Omega)$ with a score function $\scorefunction(x): \mathbb{N}^{n} \mapsto [0, 1]$. We define the function $\scorefunction$ by aggregating the medial axis of controlled smoothed versions of $\Omega$. Our formulation of $\scorefunction$ must yield high values at points $x$ that belong to the real medial axis and low values at points that belong to spurious branches. Additionally, we require $\scorefunction$ to be equivariant to isometric transformations.

\subsection{The Cosine-Pruned Medial Axis}

Let us represent $\Omega$ as a square binary image $\mathcal{I}: \mathbb{N}^{2} \mapsto \left\{ 0, 1 \right\}$ with a resolution of $M \times M$ pixels. We start the computation of the CPMA by estimating a set of smoothed versions of the $\mathcal{I}$ via the Discrete Cosine Transform (DCT) and its inverse (IDCT):

\begin{align}
    \mathfrak{F}(u, v) = \frac{1}{4}C_{u}C_{v}\sum_{x=0}^{M-1}\sum_{y=0}^{M-1} \mathcal{I} \cdot cos \left( u\pi \frac{2x+1}{2M} \right) \cdot cos \left( v\pi \frac{2y+1}{2M}\right)\label{eq:rec_DCT} \\
    \mathcal{I}(u, v) = \frac{1}{4}\sum_{u=0}^{M-1}\sum_{v=0}^{M-1} C_{u}C_{v}\mathfrak{F} \cdot cos \left( u\pi \frac{2x+1}{2M} \right) \cdot cos \left( v\pi \frac{2y+1}{2M}\right)\label{eq:rec_IDCT},
\end{align}
                
where $(u,v)$ are coordinates in the frequency domain, and $(x,y)$ are the spatial coordinates of the Euclidean space where $\Omega$ is defined. The values of $C_{u}$ and $C_{v}$ are determined by: 

\begin{displaymath}
    \arraycolsep=1.4pt\def\arraystretch{1.5}
    \begin{array}{l}
         C_{u} = \left \{ 
                    \begin{array}{ll}
                        \frac{1}{\sqrt{2}} & \textrm{if } u = 0 \\
                        1 & \textrm{otherwise} \\
                    \end{array} \right. \\
         C_{v} = (\textrm{Similar to above})  \\
    \end{array}
\end{displaymath}

The DCT is closely related to the discrete Fourier transform of real valued-functions. However, it has better energy compaction properties with only a few of the transform coefficients representing the majority of the energy. Multidimensional variants of the various DCT types follow directly from the one-dimensional definition. They are simply a separable product along each dimension. 


Let us now denote by $\mathcal{I}^{(i)}$ with $i=1, 2, ..., M$ the reconstructions of $\mathcal{I}$ using only the first $i$ frequencies as per equation \ref{eq:rec_IDCT}. We seek to obtain a $\scorefunction$ acting as a sort of probability indicating how likely it is for a point $x$ to be in the medial axis of $\Omega$. Points on relevant branches will appear regularly in the smoothed shapes' medial axis, resulting in high score function values. In contrast, spurious branches will only appear occasionally, resulting in low values.

\begin{definition}{\textbf{Score function.}} Let $\mathcal{I}: \mathbb{N}^{n} \mapsto \left\{ 0, 1 \right\}$ be a square binary image such that $\mathcal{I}(x) = 1 \; \forall x \in \Omega$. Let $\mathcal{I}^{(i)}$ also be the $i$-frequency reconstruction of $\mathcal{I}$ via the IDCT. We define $\scorefunction(x): \mathbb{N}^{n} \mapsto [0, 1]$ as the per pixel average over a set of estimations of the $\mathbf{MAT}$ on the smoothed shapes $\mathcal{I}^{(i)}$.

\begin{equation}
    \scorefunction(x) = \frac{1}{M}\sum_{i=1}^{M} \left [ \mathbf{MAT} \left ( \hat{\mathcal{I}}^{(i)} \right ) \right ](x).
    \label{def:score_function}
\end{equation}
\end{definition}

The score function is defined for all $x$ in the domain of $\mathcal{I}$. Notice that we represent $\mathbf{MAT}$ as another binary image where $\mathbf{MAT}(x) = 1$ only when $x$ belongs to the medial axis. Finally, with $\scorefunction$, we have all the elements to present our definition of the CPMA.

\begin{definition}{\textbf{Cosine-Pruned Medial Axis}} Given a binary image $\mathcal{I}: \mathbb{N}^{n} \mapsto \left\{ 0, 1 \right\}$ representing a shape $\Omega$, the $\mathbf{CPMA}(\Omega)$ consist of all the pairs $(x, r) \in \mathbf{MAT}(\Omega)$ such that $\scorefunction(x)$ is greater than a threshold $\tau$:

\begin{displaymath}
    \left[ CPMA(\Omega) \right] (x) = \left \{ 
                \begin{array}{ll}
                    1 & \scorefunction(x) > \tau\\
                    0 & \textrm{otherwise} \\
                \end{array} \right. 
\end{displaymath}
\end{definition}

\begin{figure}[!ht]
    \centering
    \begin{subfigure}[b]{\linewidth}
        \includegraphics[width=\linewidth]{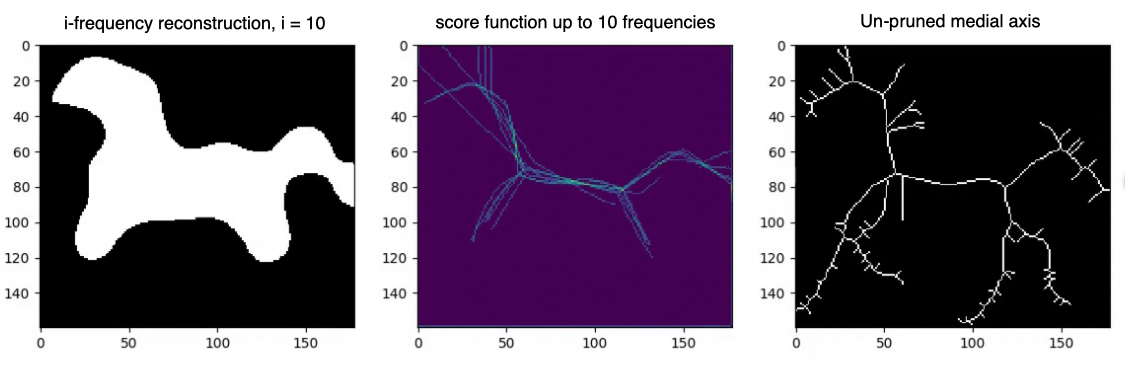}
    \end{subfigure}
    
    \begin{subfigure}[b]{\linewidth}
        \includegraphics[width=\linewidth]{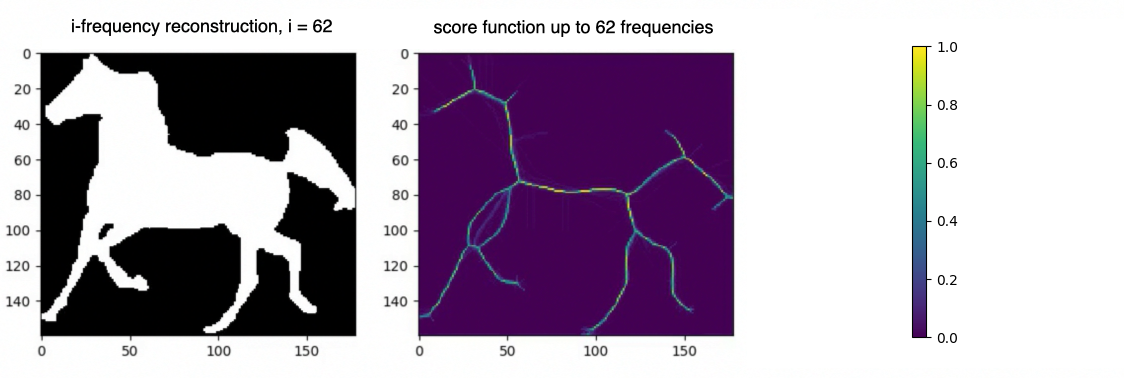}
    \end{subfigure}
    
    \caption[Score Function.]{Score Function illustrative example. The rows show $\scorefunction$ of an image of size  on a $180 \times 180$. We computed the reconstructions up to only $M_{1}=10$ and $M_{2}=62$ of the first frequencies.\label{fig:score_function}}
\end{figure}

We empirically set the value of the threshold to $\tau=0.47$. However, we conducted an additional experiment to show that this value is consistent across different shapes.

\begin{figure}[!t]
    \centering
    \includegraphics[width=0.6\linewidth]{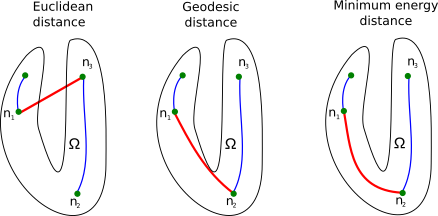}
    \caption[Path connectivity between CPMA segments]{Path connectivity between CPMA segments. When using the Euclidean distance (left), two nodes can connect through a path that is not contained within $\Omega$. The geodesic distance (center) guarantees that the path is in $\Omega$, but does not follow the center-line. The minimum energy distance (center) $E_{\Omega}$ is a better alternative to enforce the path to follow the medial axis.}
    \label{fig:cpma_connectivity}
\end{figure}

Although the CPMA results in a noise-free $\mathbf{MA}$, there is no restriction in its formulation to force the CPMA to create a connected medial axis. We solve this by finding all the disconnected pieces of the CPMA. Later, we connect them using a minimum distance criterion $g(x_{i}, x_{j})$, where $x_{i}$ and $x_{j}$ are endpoints of two distinct pieces. However, neither the Euclidean distance nor the geodesic distance are suitable criteria because they lead to connections between nodes that do not follow the medial axis (See figure \ref{fig:cpma_connectivity}). We instead connect $x_{i}$ and $x_{j}$ with a minimum energy path using an energy function $E_{\Omega}$. We must guarantee that $E_{\Omega}(x)$ has high values when $x$ is close to $\delta \Omega$ and low values when $x$ is close to the medial axis. This way, we enforce the paths to be close to the $\mathbf{MAT}$. We call the result of this strategy the Connected CPMA (C-CPMA). In section \ref{subsec:implementaion_details}, we provide details of $E_{\Omega}$ computation.


\subsection{Isometric Equivariance of the CPMA}

The distance transform-based medial axis depends only on the shape $\Omega$, not on the position or size in the embedding Euclidean space. Therefore the medial axis should be equivariant under isometric transformations.

\begin{proposition}\label{prop:MAT_equivariance} Let $\mathbf{MAT}(\Omega)$ be the medial axis transform of a connected bounded domain $\Omega$ embedded in $\mathbb{R}^{n}$, and let $R(x) = Mx + b$ be an isometric transformation in $\mathbb{R}^{n}$. The square matrix $M$ is a composition of any number or rotations and reflection matrices, and $b$ is an $n$-dimensional vector. We say that $\mathbf{MAT}(\Omega)$ is \textbf{equivariant} to any $R$ such that $\mathbf{MAT}(R(\Omega)) = R(\mathbf{MAT}(\Omega))\label{eq:MAT_equivariance}$.

\begin{proof} Recall that $R$ is an isometric transformation and thus it is invertible and preserves the Euclidean distance. The previous statement implies that $R$ is an isomorphic map between an open ball $B_{r}(x)$ and $B_{r}(R(x))$. Consequently, if $B_{r}(R(x))$ is a maximal ball in $\Omega$ then from definition \ref{def:MAT} we have that $B_{r}(R(x)) \not \nsubseteq B_{r ^{\prime}}(R(x^{\prime})), \forall r^{\prime} > r$. 

Let us now define $y = R(x)$. Applying $R$ to every element of $\mathbf{MAT}(\Omega)$, we have:



\begin{align*}
    R(\mathbf{MAT}(\Omega)) =& \left\{ (R(x), r) \;\mid\; B_{r}(x) \not \nsubseteq B_{r ^{\prime}}(x^{\prime}), \forall r^{\prime} > r  \right \} \\
     =& \left\{ (R(x), r) \mid B_{r}(R(x)) \not \nsubseteq B_{r ^{\prime}}(R(x^{\prime})), \forall r^{\prime} > r  \right \} \\
     =& \left\{ (y, r) \;\mid\; B_{r}(y) \not \nsubseteq B_{r ^{\prime}}(y^{\prime}), \forall r^{\prime} > r \right \} = \;\mathbf{MAT}(R(\Omega)).
\end{align*}

\end{proof}
\end{proposition}

Moreover, the CPMA depends primarily on $\scorefunction$, which also holds the isometric equivariant property. 

\begin{corollary} Let $\scorefunction$ be the score function of $\Omega$ as per definition \ref{def:score_function}, and let $R$ be an isometric transformation. We say that $\scorefunction$ is equivariant to any $R$ such that $R(\scorefunction) = \mathcal{F}_{R(\Omega)}$.

\begin{proof} Using the results from proposition \ref{prop:MAT_equivariance} and recalling that $R$ is a linear transformation, we have that:

\begin{displaymath}
     R(\scorefunction) = R\left(\frac{1}{M}\sum_{i=1}^{M} \mathbf{MAT}( \hat{\mathcal{I}}^{(i)})\right) 
     = \frac{1}{M}\sum_{i=1}^{M} R\left(\mathbf{MAT}( \hat{\mathcal{I}}^{(i)})\right) 
     = \frac{1}{M}\sum_{i=1}^{M} \mathbf{MAT}\left( R(\hat{\mathcal{I}}^{(i)})\right) = \mathcal{F}_{R(\Omega)}.
\end{displaymath}
\end{proof}
\end{corollary}

However, in a discrete domain, this equivariance is only an approximation because points on both $\Omega$ and $\mathbf{MAT}(\Omega)$ are constrained to be on a fixed, regular grid. In a continuous domain, it is easy to demonstrate that the cosine transform has exact isometric equivariance.




\subsection{Implementation Details}
\label{subsec:implementaion_details}


To compute the CPMA enforcing the connectivity, we create a lattice graph $\mathcal{G} = (\mathcal{V}, \mathcal{E})$. A point $p$ in the domain of  $\mathcal{I}$ is a node of $\mathcal{G}$, if and only if $p \in \Omega$. The node $p$ shares an edge with all its neighbors in the lattice only if the neighbors are also inside $\Omega$. We used an $8$-connectivity neighborhood in 2D and a $26$-connectivity neighborhood in 3D.

In order to determine the minimum energy path between pairs of pixels/voxels, we compute the minimum distance path inside $\mathcal{G}$ using Dijkstra’s algorithm. We assign weights to every edge with values extracted from $E_{\Omega}$. Given $(x, y) \in \mathcal{E}$, we compute the energy of every edge as follows:

\begin{displaymath}
     E_{\Omega}(x, y) = 1 -  \frac{\scorefunction(x) + \scorefunction(y)}{2}, \forall (x, y) \in \mathcal{E}.
\end{displaymath}

This method guarantees connectivity, but it is inefficient because of the minimum energy path's iterative computation. We sacrifice performance in favor of connectivity. We include the pseudo-code to compute the CPMA and the C-CPMA in algorithms \ref{alg:cpma} and \ref{alg:connect_skeleton_segments}.

\begin{algorithm}[!ht]
    \setstretch{1.0}
    \footnotesize
    
    \SetAlgoLined
    \DontPrintSemicolon
    
    \textbf{Input:}\;
    \hspace{0.15\linewidth} $\mathcal{I}$: N-dimensional binary array representing the object\;
    \hspace{0.15\linewidth} \textbf{M:} number of frequencies of $\mathcal{I}$ to be used in the computation\;
    \textbf{Output:}\;
    \hspace{0.15\linewidth} \textbf{CPMA:} Cosine-Pruned Medial Axis\;
    
    
    $\tau \gets  0.5$\;
    $\mathfrak{F} \gets DCT(\mathcal{I})$\;
    $\scorefunction \gets 0$\;
    
    $i \gets 1$\;
    \While{$i < M$}{
        $\hat{\mathcal{I}}^{(i)} = IDCT(\mathfrak{F}, i)$ \tcp*{Reconstructs $\mathcal{I}$ using only the first $i$ frequencies}
        $\scorefunction = \scorefunction + \mathbf{MAT}(\hat{\mathcal{I}}^{(i)})$\;
        $i \gets i + 1$\;
    }
    $\scorefunction \gets \scorefunction / M$ \tcp*{The final $\scorefunction$ is the average of all reconstructions}\;
    CPMA = $\scorefunction > \tau$\;
    \Return CPMA, $\scorefunction$\;
    
    \caption{Cosine-Pruned Medial Axis (CPMA)}
    \label{alg:cpma}
\end{algorithm}

\begin{algorithm}[!ht]
    \setstretch{1.0}
    \footnotesize 
    
    \SetAlgoLined
    \DontPrintSemicolon
    
    \textbf{Input:}\;
    \hspace{0.15\linewidth} \textbf{CPMA:} Cosine-Pruned Medial Axis representing the object\;
    \hspace{0.15\linewidth} $\scorefunction$: Score function\;
    \textbf{Output:}\;
    \hspace{0.15\linewidth} \textbf{C-CPMA:} Connected Cosine-Pruned Medial Axis\;
    
    
    C-CPMA $\gets$ \textbf{copy}(CPMA)\;
    skeleton-parts $\gets$ \textbf{compute-skeleton-parts}(CPMA)
    
    max-iter $\gets 200$\;
    it $\gets 0$\;
    \While {it $<$ max-iter \textbf{and} $|\textrm{skeleton-parts}| > 1$}{
    
        graph-i $\gets$ skeleton-parts[0]\;
        graph-f $\gets$ skeleton-parts[1]\;
        \tcp{Finds the minimum geodesic path bt. two pieces of the CPMA }
        min-path $\gets$ \textbf{find-min-path}(graph-i, graph-f, $\scorefunction$)\;
        
        C-CPMA[path] $\gets$ True\;
        
        skeleton-parts $\gets$ \textbf{compute-skeleton-parts}(C-CPMA)\;
        it $\gets$ it + $1$\;
    }
    \Return C-CPMA\;
    
    \caption{Connect skeleton segments}
    \label{alg:connect_skeleton_segments}
\end{algorithm}

The CPMA only relies on one parameter, $\tau$. The value of $\tau$ is the threshold that determines whether a point of $\scorefunction$ is a medial axis point. We empirically set the value of $\tau=0.47$. However, in section \ref{sec:results}, we present the result of an additional experiment to show how sensitive the CPMA is to different threshold values. 

Another essential consideration when computing the CPMA is the maximum frequency used to reconstruct the original shape through the IDCT. Using less than $M$ frequencies enables a faster computation of the CPMA without losing accuracy. We found that using frequencies greater than $\frac{M}{2}$ does not yield significant improvement for the CPMA. 


%% file: sections/04_experiments.tex

In this section, we describe experiments used to evaluate our approach compared to state-of-the-art medial axis pruning methods.

\subsection{Datasets}

We chose three extensively used datasets of 2D and 3D segmented objects to evaluate our methodology on medial axis extraction robustness. These datasets are part of the accepted benchmarks in literature, enabling us to compare our results.

\begin{figure*}[!ht]
    \centering
    \begin{subfigure}[b]{0.7\linewidth}
        \includegraphics[width=\linewidth]{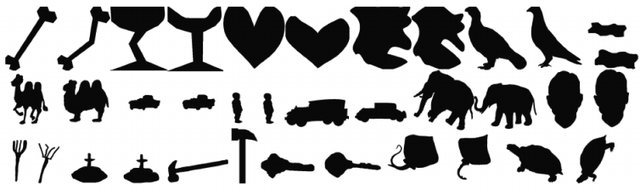}
    \caption[Kimia216 dataset.]{Kimia216 dataset.\label{fig:example_kimia_216}}
    \end{subfigure}
    
    \begin{subfigure}[b]{0.49\linewidth}
        \includegraphics[width=\linewidth]{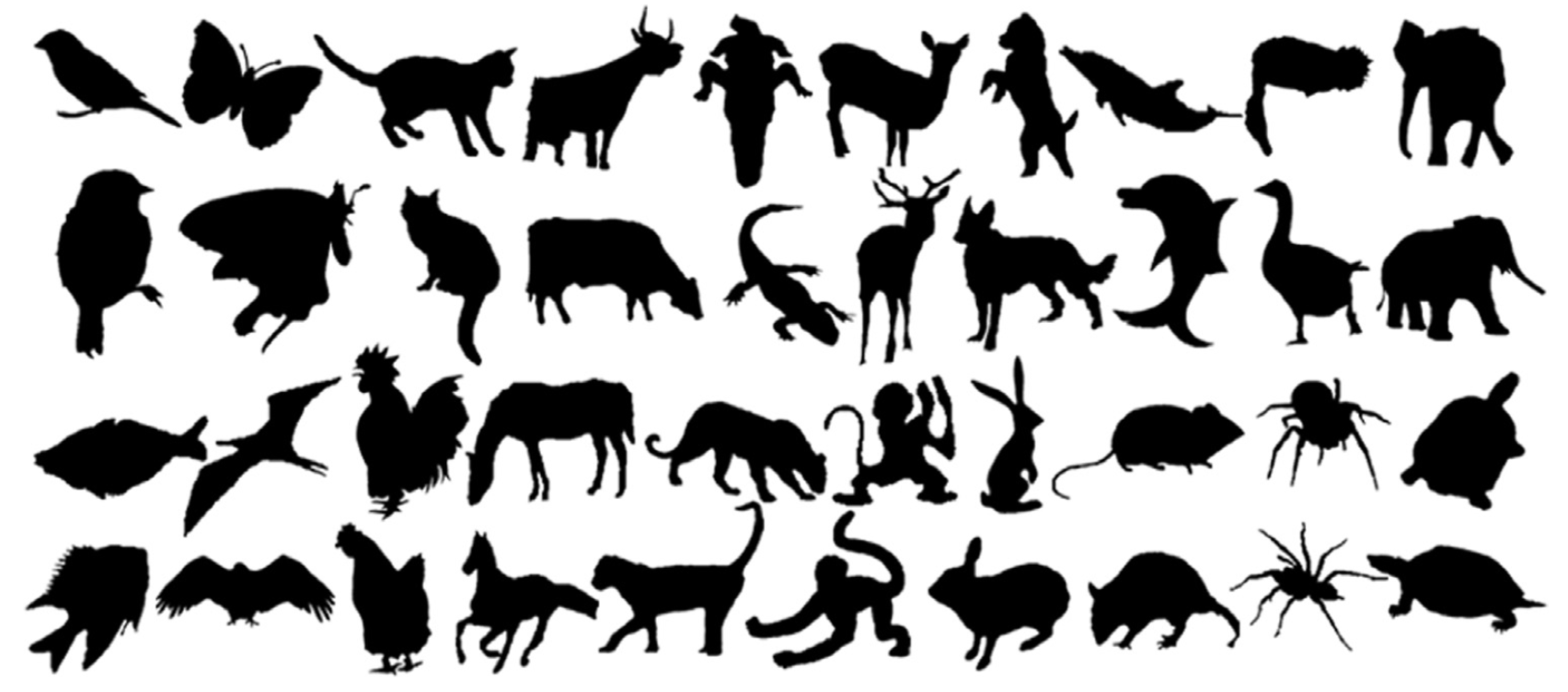}
        \caption[Animal2000 database.]{Animal2000 dataset.\label{fig:example_animal_dataset}}
    \end{subfigure}
    \begin{subfigure}[b]{0.5\linewidth}
        \includegraphics[width=\linewidth]{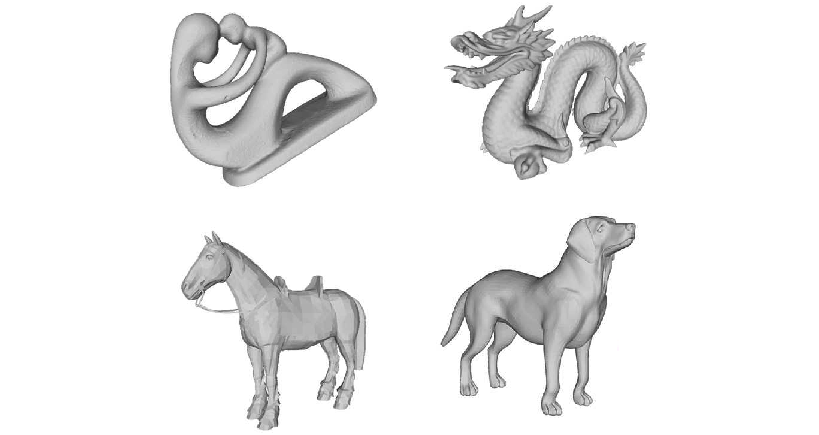}
        \caption[University of Groningen Benchmark.]{University of Groningen Benchmark.\label{fig:examples_groningen_benchmark}}
    \end{subfigure}
    \caption[Sample shapes from all the datasets used in our experiments.]{Sample shapes from all the datasets used in our experiments.}
\end{figure*}

\paragraph{\textbf{Kimia216 dataset}~\citep{Sebastian2004}} It consists of 18 classes of different shapes with 12 samples in each class. The dataset's images are a collection of slightly different views of a set of shapes with varying topology. Figure \ref{fig:example_kimia_216} shows two samples from each class. Contour noise and random rotations are also present in some images in the dataset. Kimia216 has been largely used to test a wide range of medial axis extraction algorithms. Because of the large variety of shapes, we assume that this benchmark ensured a fair comparison with the state-of-the-art. 


\paragraph{\textbf{Animal2000 dataset}~\citep{Bai2009}} The Animal2000 dataset helps us to evaluate the properties of our approach in the presence of non-rigid transformations. It contains 2000 silhouettes of animals in 20 categories. Each category consists of 100 images of the same type of animal in different poses (Figure \ref{fig:example_animal_dataset}). Because silhouettes in Animal2000 come from real images, each class is characterized by a large intra-class variation.


\paragraph{\textbf{University of Groningen's Benchmark}} This set of 3D meshes is commonly found in the literature to evaluate medial axis extraction methods in 3D~\citep{Sobiecki2014,Sobiecki2013,Chaussard2011}. It includes scanned and synthetic shapes taken from other popular datasets. It contains shapes with and without holes, shapes of varying thickness, and shapes with smooth and noisy boundaries. See Figure \ref{fig:examples_groningen_benchmark}. All meshes are pre-processed, ensuring a consistent orientation, closeness, non-duplicated vertices, and no degenerate faces. 


\subsection{Sensitivity to Noise and Equivariance Analysis}\label{subsec:sens_to_noise_equivariance}

To compare the robustness of a medial axis extraction method, we adopt an evaluation strategy similar to \citep{Chaussard2011}. Consequently, we measure the similarity between the medial axis of a shape $\Omega$ and $\Omega^{\prime}$. The shape $\Omega^{\prime}$ derives from a ``perturbation'' of $\Omega$. We are interested in evaluating how well our methodology responds to induced noise on the contour/surface. We are also interested in assessing how stable the CPMA responds in the presence of rotations of $\Omega$ to test its isometric equivariance.

We employ the Hausdorff distance ($d_{H}$), and Dubuisson-Jain dissimilarity ($d_{D}$) as metrics between shapes. The Dubuisson-Jain similarity is a normalization of the Hausdorff distance~\citep{Dubuisson2002}, which aims to overcome $d_{H}$ sensitivity to outliers. The Dubuisson-Jain similarity between point sets $X$ and $Y$ is defined as:

\begin{equation}
d_{D}(X, Y) = max \left \{ D(X|Y), D(Y|X) \right \},
\label{eq:dubuison_jain_disimilarity}
\end{equation}
with 
\begin{equation}
D(X|Y) = \frac{1}{|X|}\sum_{x \in X} \min _{y \in Y} \left \{ d(x, y) \right\}.
\end{equation}

We must first choose a strategy to induce noise to the boundary to evaluate the noise sensitivity. We use a stochastic approach denoted by $\mathcal{E}$, where a random number of points $p \in \delta \Omega$ are deformed by a vector $v$ in the direction orthogonal to the boundary, with a deformation magnitude that is normally distributed, $|v| \sim \mathcal{N}(p, 1)$.  This noise model is recursively applied $n$ times to every shape in our datasets. We denote as $\mathbf{MAT}(\mathcal{E}(\Omega, k))$ the medial axis of a shape $\Omega$ after applying the noise model $k$ times. In our experiments, we used $k=1,...,20$ noise levels.


For every object in each dataset, we compared the medial axis $\mathbf{MAT}(\Omega)$ with the noisy versions $\mathbf{MAT}(\mathcal{E}(\Omega, k))$ to determine how sensitive a method is to boundary noise. 

The medial axis is ideally an isometric equivariant shape representation so that $\textrm{MA}(R(\Omega)) = R(\textrm{MA}(\Omega))$. Due to sampling factors, this relationship is only an approximation. However, we can measure the equivariance by comparing $\mathbf{MAT}(R(\Omega))$ with $R(\mathbf{MAT}(\Omega))$. The more similar they are, the more equivariant the method.

Because the translation equivariance is trivial, we evaluate isometric equivariance only with rotations of $\Omega$. We do not use reflections because the properties of rotation matrices we use in this study can be extended to reflection matrices. In our experiments, 2D rotations are counterclockwise in the range $[0, 90]$ degrees around the origin. We use $30$ rotations at regular intervals. In 3D, we use a combination of azimuthal ($\theta \in [0, 90]$) and elevation ($\phi \in [0, 90]$) rotations around the origin. The angles $\theta$ and $\phi$ take values each $18$ degrees. 

\subsection{Comparative Studies}

We chose seven of the most relevant methods in the scientific literature to compare with CPMA extraction results. Each method was selected based on a careful review of the state-of-the-art. These methods illustrate the variety of approaches that authors employ to prune the medial axis. Notice that the first method we used in our study is the un-pruned $\mathbf{MAT}$.

Table \ref{tab:skeletonization_comparative_study} summarizes all of the methods in our comparative study. In many cases, the performance of a pruning method depended on its parametrization. We selected parameters for all of the methods following the best performance parametrization reported in the state-of-the-art.

\begin{table}[!ht]
    \centering
    \footnotesize
    \caption{Pruning methods employed for the comparative study in 2D\label{tab:skeletonization_comparative_study}}
    
    \begin{tabular}{clp{0.3\linewidth}p{0.35\linewidth}}
        \hline
        \textbf{{Dim.}} & \textbf{{Method}} & \textbf{{Full name}} & \textbf{Parameter description} \\
        \hline
        \multirow{7}{*}{2D} & MAT~\cite{Blum1967} & Medial Axis Transform & N/A \\
         & Thinning~\cite{Zhang1984} & Zhang-Suen Algorithm & N/A \\
         & GIMA~\cite{Hesselink2008} & Gamma Integer Medial Axis & $\mathbf{\gamma}$: minimum distance between $\Pi_{\Omega}(x)$ and $\Pi_{\Omega}(y)$, $y \in N_{x}$, the neighborhood of $x$. \\
         & BEMA~\cite{Couprie2007} & Bisector Euclidean Medial Axis & $\mathbf{\theta}$: angle formed by the point $x$ and the two projections $\Pi_{\Omega}(x)$ and $\Pi_{\Omega}(y)$, $y \in N_{x}$.\\
         & SAT~\cite{Giesen2009} & Scale Axis Transform & $\mathbf{s}$: scale factor to resize $\textbf{MAT}(\Omega)$.  \\
         & SFEMA~\cite{Postolski2014} & Scale Filtered Euclidean Medial Axis & $\mathbf{s}$: scale factor to all balls in the $\textbf{MAT}(\Omega)$. \\
         & Poisson skel.~\cite{Aubert2014} & Poisson Skeleton & $\mathbf{w}$: window size to find the local maximum of contour curvature.\\
        \hline
        \multirow{2}{*}{3D} & Thinning~\citep{Zhang1984} & Zhang-Suen Algorithm & N/A \\
         & TEASAR~\citep{Sato2000} & Tree-structure skeleton extraction & N/A \\
        \hline
    \end{tabular}
    \caption*{This table shows name and parameter description for each method. The point $x \in \textbf{MAT}$ is an element of the MAT that can be potentially pruned. $\Pi_{\Omega}(x)$ refers to the set of closest boundary points of $x$. All the elements of $\Pi_{\Omega}(x)$ have the same distance from $x$.}
\end{table}

%% file: sections/05_results.tex
\label{sec:results}

In this section, we present and discuss the results of our approach on medial axis pruning. We focus on two properties: 1) robustness to noise of the contour and 2) isometric equivariance.

\subsection{Stability Under Boundary Noise}

We compared the stability of the CPMA under boundary noise against other approaches in Table \ref{tab:skeletonization_comparative_study}. We conducted our experiments on Kimia216 and the Animal2000 Dataset for 2D images. Additionally, we used a set of three-dimensional triangular meshes from the Groningen Benchmark for 3D experimentation.

For our noise sensitivity experiments, we applied $20$ times the noise model $\mathcal{E}(\Omega, k)$ to every object of each dataset. We then computed their $\mathbf{MAT}$ using every method in Table \ref{tab:skeletonization_comparative_study} with different parameters. Finally, each $\mathbf{MAT}(E(\Omega, k))$ was compared with $\mathbf{MAT}(\Omega)$ using both the Hausdorff distance and Dubuisson-Jain dissimilarity. We report the per method average of each metric over all the elements of each dataset.

\begin{figure*}[!ht]
    \centering
    \begin{subfigure}[b]{0.410\linewidth}
        \includegraphics[width=\linewidth]{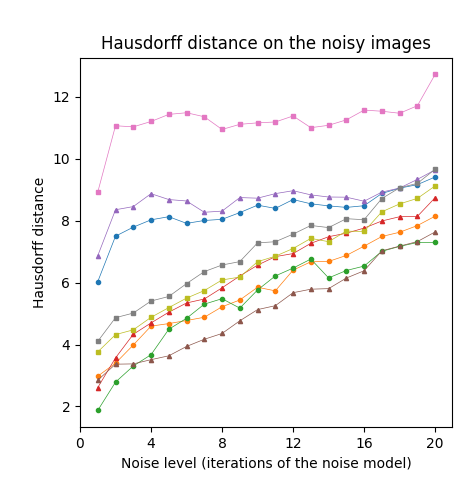}
    \end{subfigure}
    \begin{subfigure}[b]{0.580\linewidth}
        \includegraphics[width=\linewidth]{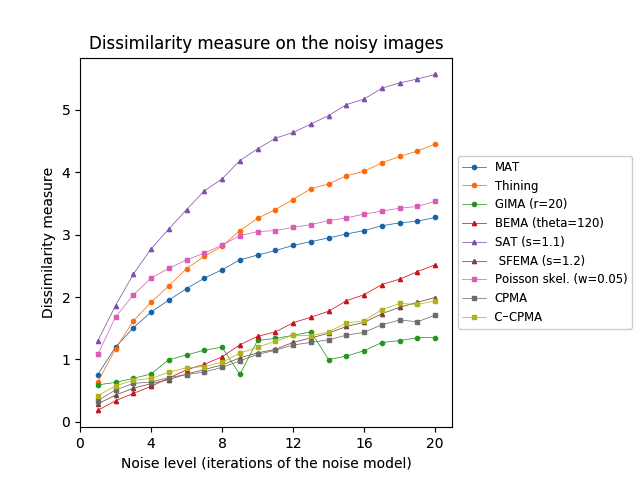}
    \end{subfigure}
    
    \begin{subfigure}[b]{0.410\linewidth}
        \includegraphics[width=\linewidth]{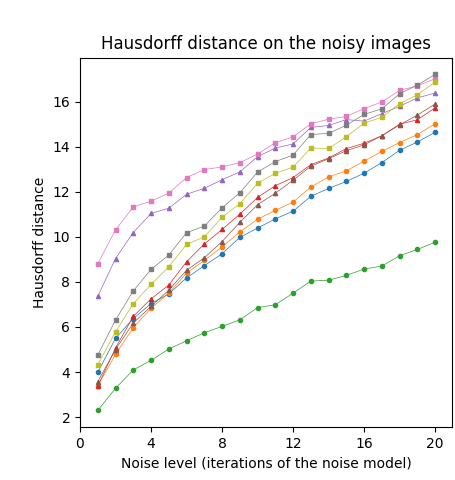}
    \end{subfigure}
    \begin{subfigure}[b]{0.580\linewidth}
        \includegraphics[width=\linewidth]{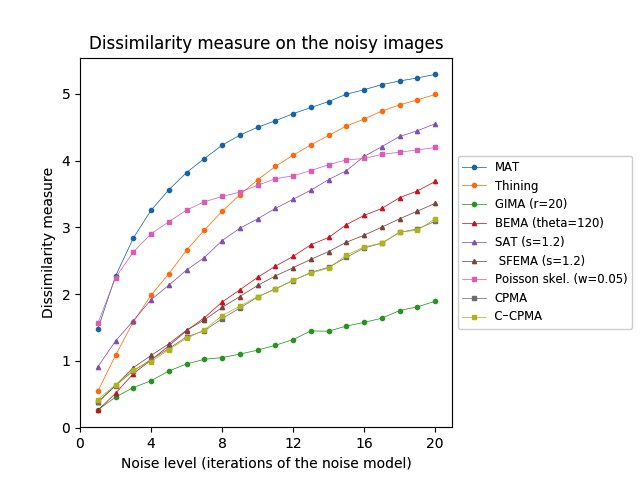}
    \end{subfigure}
    
    \begin{subfigure}[b]{0.410\linewidth}
        \includegraphics[width=\linewidth]{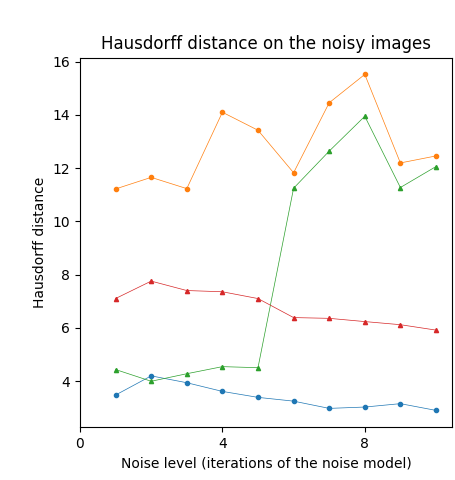}
    \end{subfigure}
    \begin{subfigure}[b]{0.580\linewidth}
        \includegraphics[width=\linewidth]{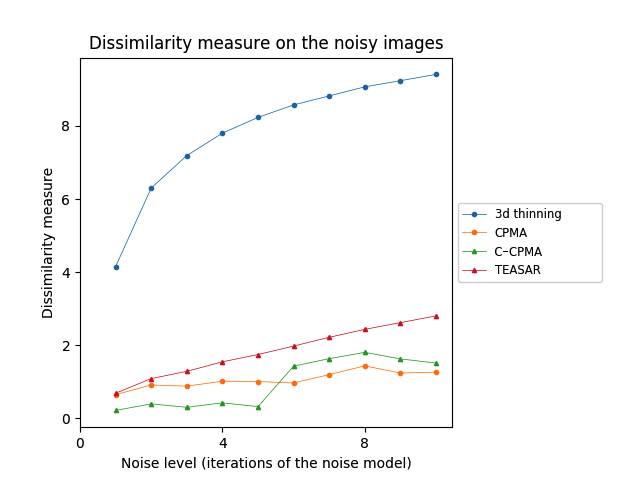}
    \end{subfigure}
    
    \caption[Noise sensitivity results on Kimia216 dataset (top), Animal2000 dataset (middle), and Groningen Benchmark (bottom)]{Noise sensitivity results on Kimia216 dataset (top), Animal2000 dataset (middle), and Groningen Benchmark (bottom). The figure shows the Hausdorff distance (left) and the Dubuisson-Jain dissimilarity (right) for all of the methods in Tables \ref{tab:noise_sensitivity_results_kimia}, \ref{tab:noise_sensitivity_results_animal}, and \ref{tab:noise_sensitivity_results_groningen}. Only the best parametrization of each method is depicted for better interpretation.\label{fig:noise_results}}
\end{figure*}

First, we tested our medial axis pruning method on the Kimia216 dataset and present the results in Table \ref{tab:noise_sensitivity_results_kimia}. The table shows that the CPMA and the C-CPMA are competitive against state-of-the-art methods for medial axis extraction. Our results show similar performance to two state-of-the-art methods: the GIMA and SFEMA. The CPMA and C-CPMA also performed better than Poisson Skeletons, SAT, topological thinning, and the MAT itself. For visual comparison, Figure \ref{fig:noise_results} (top row) shows both Hausdorff distance and Dubuisson-Jain dissimilarity against noise level. The figure only displays the best parametrization of every method to improve visualization. It is clear that the CPMA and the C-CPMA are among the three best results when we use the Dubuisson-Jain dissimilarity metric. However, we observe a decrease in performance when we the Hausdorff distance metric. We believe this occurs because of Hausdorff's distance sensitivity to outliers.

\begin{table}[!ht]
    \centering
    \footnotesize
    \caption{Noise sensitivity results on Kimia216.\label{tab:noise_sensitivity_results_kimia}}
    
    \begin{tabular}{l|cccc|cccc}
        \hline
        \textbf{{Method}} & \multicolumn{4}{c}{\textbf{{Hausdorff}}} & \multicolumn{4}{c}{\textbf{{Dissimilarity}}} \\
          & 5 & 10 & 15 & 20  & 5 & 10 & 15 & 20 \\
        \hline
        MAT  & 8.13 & 8.50 & 8.43 & 9.41  & 1.95 & 2.67 & 3.01 & 3.27 \\
        Thinning  & 4.68 & 5.85 & 6.88 & 8.15  & 2.18 & 3.26 & 3.94 & 4.45 \\
        GIMA (r=5)  & 5.46 & 6.50 & 7.37 & 8.84  & 0.87 & 1.31 & 1.60 & 1.88 \\
        GIMA (r=10)  & 5.40 & 7.12 & 8.35 & 9.18  & \textbf{0.68} & 1.08 & 1.35 & 1.58 \\
        GIMA (r=20)  & 4.49 & 5.76 & 6.39 & \textbf{7.30}  & 1.00 & 1.30 & \textbf{1.05} & \textbf{1.35} \\
        BEMA (theta=90)  & 5.22 & 6.55 & 7.11 & 8.30  & 0.99 & 1.60 & 2.07 & 2.53 \\
        BEMA (theta=120)  & 5.05 & 6.56 & 7.60 & 8.74  & 0.70 & 1.37 & 1.94 & 2.52 \\
        BEMA (theta=150)  & 6.68 & 7.69 & 7.89 & 9.40  & 0.99 & 1.80 & 2.50 & 3.37 \\
        SAT (s=1.1)  & 8.68 & 8.73 & 8.76 & 9.64  & 3.09 & 4.37 & 5.08 & 5.57 \\
        SAT (s=1.2)  & 9.61 & 10.05 & 9.79 & 10.20  & 2.50 & 3.22 & 3.89 & 4.47 \\
         SFEMA (s=1.1)  & 4.15 & 5.35 & 6.18 & 7.53  & 0.84 & 1.37 & 1.92 & 2.50 \\
         SFEMA (s=1.2)  & \textbf{3.64} & \textbf{5.13} & \textbf{6.15} & 7.64  & 0.68 & 1.11 & 1.53 & 1.99 \\
        Poisson skel. (w=0.05)  & 11.43 & 11.16 & 11.26 & 12.73  & 2.46 & 3.05 & 3.27 & 3.53 \\
        Poisson skel. (w=0.10)  & 15.60 & 15.48 & 16.07 & 17.35  & 3.62 & 4.07 & 4.19 & 4.56 \\
        Poisson skel. (w=0.20)  & 17.71 & 18.02 & 19.54 & 21.35  & 5.00 & 5.38 & 5.63 & 6.20 \\
        \hline
        CPMA  & 5.55 & 7.28 & 8.07 & 9.66  & 0.71 & \textbf{1.07} & 1.39 & 1.71 \\
        C-CPMA  & 5.19 & 6.68 & 7.66 & 9.12  & 0.80 & 1.20 & 1.58 & 1.94 \\
        \hline
    \end{tabular}
    \caption*{The table shows the average Hausdorff distance and Dubuisson-Jain dissimilarity for different noise levels (5-20) over each element of the dataset.}
\end{table}

The Animal2000 dataset contains nearly ten times more shapes than Kimia216. This leads to more variation between shapes, and therefore a more challenging setting. Table \ref{tab:noise_sensitivity_results_animal} shows similar results compared to Kimia216, confirming that the noise invariant properties of the CPMA still hold in a more robust dataset. The GIMA and the SFEMA are still the best methods measured with the Dubuisson-Jain dissimilarity, closely followed by both the CPMA and the C-CPMA. Results of using the Dubuisson-Jain dissimilarity as a metric show that the CPMA is close to methods such as BEMA and SFEMA. However, the results are not as good as when using the Hausdorff distance metric. Figure \ref{fig:noise_results} (middle row) depicts the best performance for every method in comparison to ours. Our experiment's results suggest that the CPMA noise invariant properties generalize across different datasets.

\begin{table}[!ht]
    \centering
    \footnotesize
    \caption{Noise sensitivity results on Animal2000.\label{tab:noise_sensitivity_results_animal}}
    
    \begin{tabular}{l|cccc|cccc}
        \hline
        \textbf{{Method}} & \multicolumn{4}{c}{\textbf{{Hausdorff}}} & \multicolumn{4}{c}{\textbf{{Dissimilarity}}} \\
          & 5 & 10 & 15 & 20  & 5 & 10 & 15 & 20 \\
        \hline
        MAT  & 7.47 & 10.39 & 12.47 & 14.64  & 3.56 & 4.50 & 5.00 & 5.29 \\
        Thinning  & 7.51 & 10.79 & 12.94 & 15.03  & 2.30 & 3.71 & 4.52 & 4.99 \\
        GIMA (r=5)  & 7.96 & 11.00 & 13.23 & 15.27  & 1.24 & 1.88 & 2.35 & 2.68 \\
        GIMA (r=10)  & 6.78 & 8.56 & 10.21 & 11.54  & 0.89 & 1.29 & 1.62 & 1.89 \\
        GIMA (r=20)  & \textbf{5.02} & \textbf{6.86} & \textbf{8.29} & \textbf{9.77}  & \textbf{0.85} & \textbf{1.16} & \textbf{1.52} & \textbf{1.89} \\
        BEMA (theta=90)  & 7.84 & 10.61 & 12.76 & 14.78  & 1.45 & 2.37 & 3.06 & 3.57 \\
        BEMA (theta=120)  & 7.86 & 11.76 & 13.93 & 15.74  & 1.22 & 2.25 & 3.04 & 3.69 \\
        BEMA (theta=150)  & 8.88 & 12.38 & 14.39 & 16.51  & 1.68 & 3.00 & 3.95 & 4.72 \\
        SAT (s=1.1)  & 9.44 & 11.80 & 13.47 & 15.21  & 2.80 & 4.13 & 5.02 & 5.49 \\
        SAT (s=1.2)  & 11.28 & 13.57 & 15.21 & 16.40  & 2.13 & 3.13 & 3.85 & 4.55 \\
         SFEMA (s=1.1)  & 7.44 & 11.00 & 13.30 & 15.35  & 1.33 & 2.27 & 3.03 & 3.69 \\
         SFEMA (s=1.2)  & 7.64 & 11.43 & 13.83 & 15.90  & 1.26 & 2.13 & 2.78 & 3.36 \\
        Poisson skel. (w=0.05)  & 11.94 & 13.68 & 15.35 & 17.03  & 3.08 & 3.63 & 4.01 & 4.20 \\
        Poisson skel. (w=0.10)  & 14.55 & 17.11 & 18.64 & 20.10  & 3.55 & 4.08 & 4.68 & 4.94 \\
        Poisson skel. (w=0.20)  & 17.37 & 20.35 & 21.67 & 23.69  & 3.94 & 4.69 & 5.33 & 5.73 \\
        \hline
        CPMA  & 9.20 & 12.88 & 14.96 & 17.22  & 1.18 & 1.96 & 2.55 & 3.09 \\
        C-CPMA  & 8.67 & 12.39 & 14.45 & 16.88  & 1.17 & 1.96 & 2.58 & 3.13 \\
        \hline
    \end{tabular}
    \caption*{Noise sensitivity results on Animal2000. The table shows the average Hausdorff distance and Dubuisson-Jain dissimilarity for different noise levels (5-20) over each element of the dataset.}
\end{table}

    

For our 3D experiments, we selected $14$ objects from the Groningen Benchmark, reflecting the most common shapes used in the literature. Each object was voxelized to a binary voxel grid with resolution $150 \times 150 \times 150$. This resolution offered sufficient details as well as a sufficiently low computational cost. In contrast to the 2D case, we applied $\mathcal{E}(\Omega, k)$ only $10$ times to the 3D object. We did this for two reasons: 1) to reduce computational complexity and 2) noise tends to be more extreme in 3D at the chosen resolution. The results on the Groningen dataset are shown in Table \ref{tab:noise_sensitivity_results_groningen} and Figure \ref{fig:noise_results} (bottom row). Notice that both the CPMA and C-CPMA achieved the best results among the other methods when compared with the dissimilarity measure. These results show that our methodology has noise-invariance properties, and it is stable in the presence of small surface deformation. However, the results show unusual patterns when compared with the Hausdorff distance. In fact, for some methods, the metric decreases when the noise level increases. We attribute this behavior to the outlier sensibility of the Hausdorff distance.

\begin{table}[!ht]
    \centering
    \tabcolsep 5pt
    \footnotesize
    \caption{Noise sensitivity results on Groningen Benchmark.\label{tab:noise_sensitivity_results_groningen}}
    
    \begin{tabular}{l|ccccc|ccccc}
        \hline
        \textbf{{Method}} & \multicolumn{5}{c}{\textbf{{Hausdorff}}} & \multicolumn{4}{c}{\textbf{{Dissimilarity}}} \\
          & 2 & 4 & 6 & 8 & 10  & 2 & 4 & 6 & 8 & 10 \\
        \hline
        3D thinning  & 4.20 & \textbf{3.61} & \textbf{3.25} & \textbf{3.03} & \textbf{2.90}  & 6.30 & 7.80 & 8.58 & 9.07 & 9.41 \\
        TEASAR  & 7.76 & 7.36 & 6.39 & 6.24 & 5.92  & 1.09 & 1.55 & 1.98 & 2.44 & 2.80 \\
        \hline
        CPMA  & 11.66 & 14.10 & 11.83 & 15.52 & 12.46  & 0.91 & 1.02 & \textbf{0.97} & \textbf{1.44} & \textbf{1.27} \\
        C-CPMA& \textbf{3.99} & 4.54 & 11.25 & 13.95 & 12.06  & \textbf{0.40} & \textbf{0.43} & 1.43 & 1.81 & 1.51 \\
        \hline
    \end{tabular}
    \caption*{The table shows the average Hausdorff distance and Dubuisson-Jain dissimilarity for different noise levels (5-20) over each element of the dataset. }
\end{table}

    

\begin{figure*}[!ht]
    \centering
    \begin{subfigure}[b]{1.0\linewidth}
        \includegraphics[width=\linewidth]{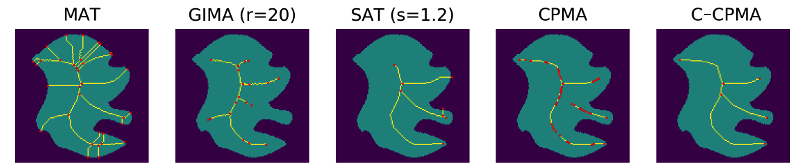}
    \end{subfigure}
    
    \begin{subfigure}[b]{1.0\linewidth}
        \includegraphics[width=\linewidth]{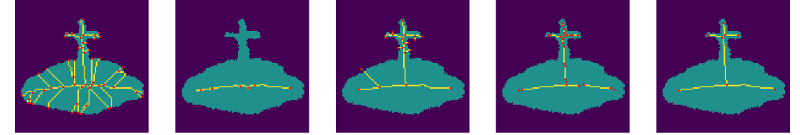}
    \end{subfigure}
    
    \begin{subfigure}[b]{1.0\linewidth}
        \includegraphics[width=\linewidth]{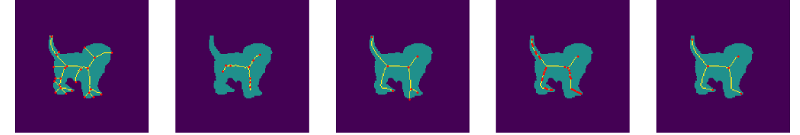}
    \end{subfigure}
    
    \begin{subfigure}[b]{1.0\linewidth}
        \includegraphics[width=\linewidth]{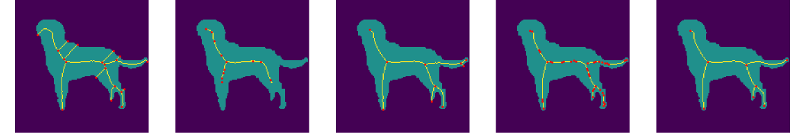}
    \end{subfigure}
    
    \begin{subfigure}[b]{1.0\linewidth}
        \includegraphics[width=\linewidth]{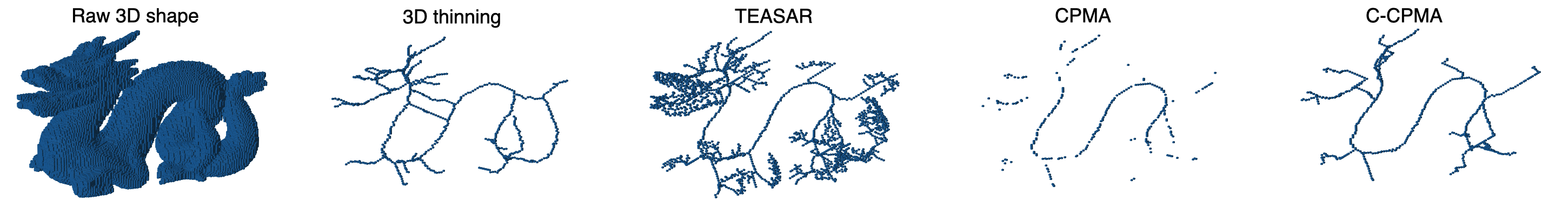}
    \end{subfigure}
    
    \begin{subfigure}[b]{1.0\linewidth}
        \includegraphics[width=\linewidth]{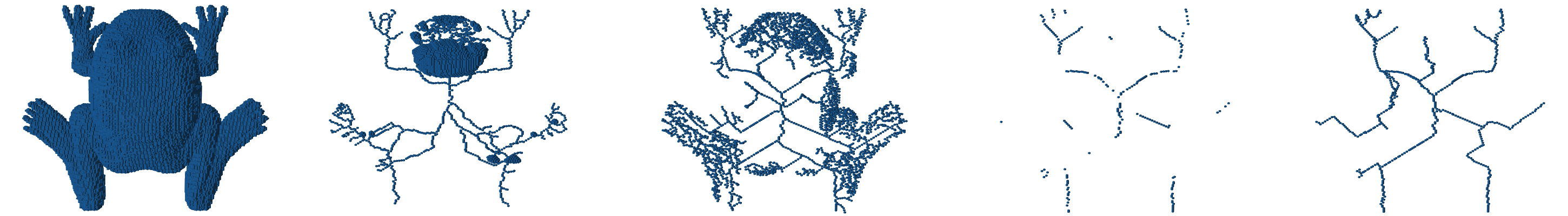}
    \end{subfigure}
    
    \caption{The images show the un-pruned $\textrm{MAT}$ and the results of four different pruning methods. Rows one and two are objects from Kimia216 dataset. Rows three and four are objects from Animal2000. Rows five and six are objects from the Groningen Benchmark. Notice how the CPMA and the C-CPMA yield medial axes with less spurious branches while preserving the topology.\label{fig:mat_examples_noise}}
\end{figure*}

We complete the noise stability analysis showing examples of the $\mathbf{MAT}$ computed with our methodology, and compared to the other methods in this study, Figure \ref{fig:mat_examples_noise}.

\subsection{Sensitivity to Rotations}

\begin{figure*}[!ht]
    \centering
    \begin{subfigure}[b]{0.410\linewidth}
        \includegraphics[width=\linewidth]{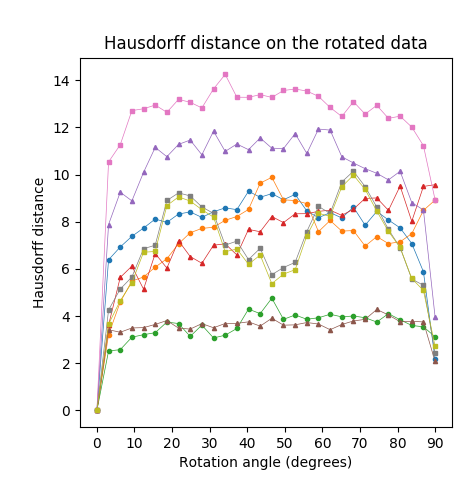}
    \end{subfigure}
    \begin{subfigure}[b]{0.580\linewidth}
        \includegraphics[width=\linewidth]{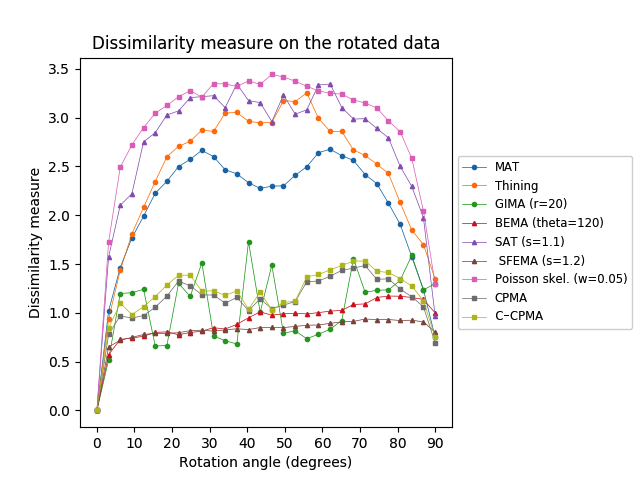}
    \end{subfigure}
    
    \begin{subfigure}[b]{0.410\linewidth}
        \includegraphics[width=\linewidth]{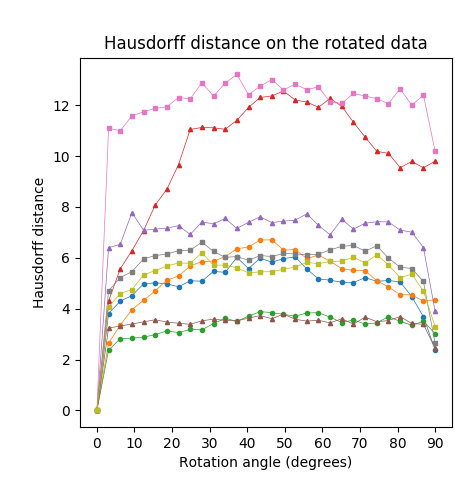}
    \end{subfigure}
    \begin{subfigure}[b]{0.580\linewidth}
        \includegraphics[width=\linewidth]{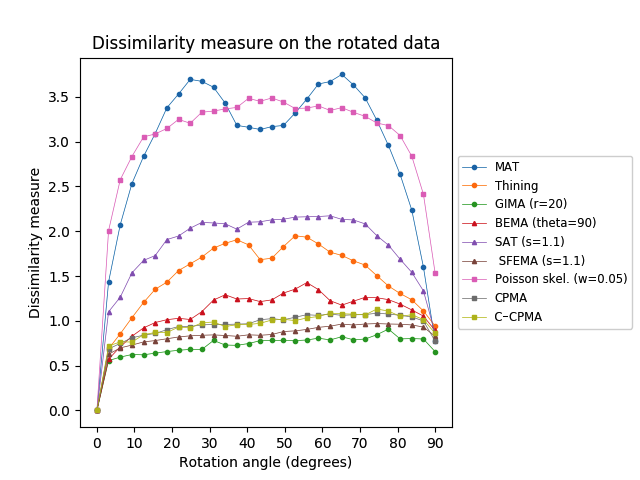}
    \end{subfigure}
    
    \begin{subfigure}[b]{0.384\linewidth}
        \includegraphics[width=\linewidth]{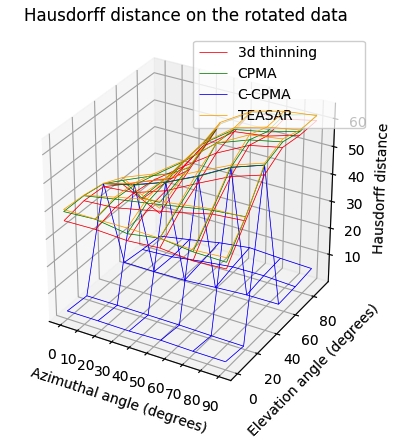}
    \end{subfigure}
    \begin{subfigure}[b]{0.384\linewidth}
        \includegraphics[width=\linewidth]{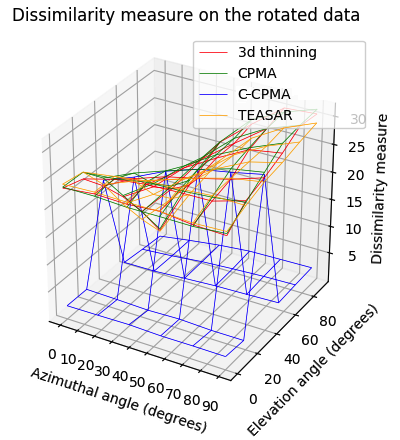}
    \end{subfigure}
    
    \caption[Rotation equivariance results on Kimia216 dataset (top), Animal2000 (middle), and Groningen Benchmark (bottom)]{Rotation equivariance results on Kimia216 dataset (top), Animal2000 (middle), and Groningen Benchmark (bottom). The top row shows the Hausdorff distance and Dubuisson-Jain dissimilarity for all the methods in Tables \ref{tab:rotation_sensitivity_results_kimia} and \ref{tab:rotation_sensitivity_results_animal}.\label{fig:rotation_results}}
\end{figure*}

We measured the dissimilarity between $\mathbf{MAT}(R(\Omega))$ and $R(\mathbf{MAT}(\Omega))$ for different instances, and different definitions of the medial axis transform. The lower this dissimilarity, the more stable the method is under rotation. The rotation sensitivity analysis on the Kimia216 dataset is summarized in Table \ref{tab:rotation_sensitivity_results_kimia} and Figure \ref{fig:rotation_results} (top row). The results show that the curves of the CPMA and the C-CPMA fall near the average of the rest of the methods achieving state-of-the-art performance. The results also surpassed several methods, such as Poisson skeleton, SAT, and thinning methods. Notice that when using the dissimilarity metric the CPMA, the GIMA, the SFEMA, and the BEMA form a subgroup that performs significantly better compared to the others. Moreover, the performance of these methods oscillates around a value of dissimilarity of around $1$ pixel on average. The intuition for this result is that regardless of the rotation, skeletons computed with these methods vary only at one pixel. Consequently, we can claim that they exhibit rotation equivariance.

\begin{table}[!ht]
    \centering
    \footnotesize
    \caption{Rotation equivariance results on Kimia216.\label{tab:rotation_sensitivity_results_kimia}}
    
    \begin{tabular}{l|ccc|ccc}
        \hline
        \textbf{{Method}} & \multicolumn{3}{c}{\textbf{{Hausdorff}}} & \multicolumn{3}{c}{\textbf{{Dissimilarity}}} \\
          & 30º & 60º & 90º  & 30º & 60º & 90º \\
        \hline
        MAT  & 8.18 & 8.17 & 2.17  & 2.67 & 2.64 & 0.75 \\
        Thinning  & 7.72 & 7.58 & 8.92  & 2.87 & 2.99 & 1.35 \\
        GIMA (r=5)  & 6.16 & 6.03 & 5.54  & 1.02 & 1.11 & 0.85 \\
        GIMA (r=10)  & 5.54 & 6.25 & 5.04  & 0.83 & 1.02 & 0.72 \\
        GIMA (r=20)  & \textbf{3.62} & 3.93 & 3.12  & 1.51 & \textbf{0.78} & 1.30 \\
        BEMA (theta=90)  & 12.35 & 12.76 & 11.72  & 1.31 & 1.60 & 1.06 \\
        BEMA (theta=120)  & 6.24 & 8.44 & 9.57  & 0.81 & 1.00 & 1.00 \\
        BEMA (theta=150)  & 9.14 & 10.09 & 10.27  & 1.11 & 1.35 & 1.36 \\
        SAT (s=1.1)  & 10.84 & 11.93 & 3.96  & 3.22 & 3.34 & 0.97 \\
        SAT (s=1.2)  & 11.44 & 12.40 & 4.45  & 2.64 & 2.87 & 0.97 \\
        SFEMA (s=1.1)  & 3.86 & 3.98 & 2.52  & 0.92 & 1.01 & 0.83 \\
        SFEMA (s=1.2)  & 3.68 & \textbf{3.66} & \textbf{2.08}  & \textbf{0.82} & 0.88 & 0.80 \\
        Poisson skel. (w=0.05)  & 12.83 & 13.32 & 8.93  & 3.21 & 3.28 & 1.30 \\
        Poisson skel. (w=0.10)  & 16.36 & 17.03 & 10.17  & 4.24 & 4.30 & 1.78 \\
        Poisson skel. (w=0.20)  & 18.94 & 19.90 & 9.65  & 5.61 & 5.66 & 2.69 \\
        \hline
        CPMA  & 8.63 & 8.65 & 2.42  & 1.18 & 1.33 & \textbf{0.70} \\
        C-CPMA  & 8.51 & 8.36 & 2.72  & 1.22 & 1.39 & 0.75 \\
        \hline
    \end{tabular}
    \caption*{Rotation equivariance results on Kimia216. The table shows the average Hausdorff distance and Dubuisson-Jain dissimilarity for different rotations of each element in the dataset.}
\end{table}

We applied the same analysis to the Animal2000 dataset achieving similar results. In this case, the CPMA and the C-CPMA ranked third and fourth, respectively, among all methods when we used the dissimilarity metric. The results for all methods and parameters are presented in Table \ref{tab:rotation_sensitivity_results_animal}. As before, we also present a summary with the best parametrization for each method in Figure \ref{fig:rotation_results} (middle row) to facilitate the interpretation. Notice that due to the larger number of objects in the Animal2000 dataset, the curves for every method appear to be smoother, highlighting stability across different rotation angles and shapes.

\begin{table}[!ht]
    \centering
    \footnotesize
    \caption{Rotation equivariance results on Animal2000.\label{tab:rotation_sensitivity_results_animal}}
    
    \begin{tabular}{l|ccc|ccc}
        \hline
        \textbf{{Method}} & \multicolumn{3}{c}{\textbf{{Hausdorff}}} & \multicolumn{3}{c}{\textbf{{Dissimilarity}}} \\
          & 30º & 60º & 90º  & 30º & 60º & 90º \\
        \hline
        MAT  & 5.08 & 5.17 & 2.38  & 3.67 & 3.64 & 0.77 \\
        Thinning  & 5.85 & 6.11 & 4.35  & 1.71 & 1.86 & 0.94 \\
        GIMA (r=5)  & 5.58 & 5.54 & 5.62  & 0.96 & 1.08 & 0.83 \\
        GIMA (r=10)  & 5.42 & 5.50 & 4.29  & 0.78 & 0.88 & 0.74 \\
        GIMA (r=20)  & \textbf{3.17} & 3.84 & 3.02  & \textbf{0.68} & \textbf{0.81} & \textbf{0.66} \\
        BEMA (theta=90)  & 11.12 & 11.92 & 9.80  & 1.10 & 1.35 & 0.89 \\
        BEMA (theta=120)  & 5.45 & 6.10 & 6.80  & 0.77 & 0.90 & 0.86 \\
        BEMA (theta=150)  & 7.60 & 8.88 & 9.91  & 1.13 & 1.36 & 1.40 \\
        SAT (s=1.1)  & 7.41 & 7.27 & 3.90  & 2.10 & 2.16 & 0.86 \\
        SAT (s=1.2)  & 9.82 & 9.62 & 4.85  & 1.77 & 1.90 & 0.95 \\
         SFEMA (s=1.1)  & 3.52 & \textbf{3.54} & 2.45  & 0.84 & 0.93 & 0.83 \\
         SFEMA (s=1.2)  & 3.54 & 3.63 & \textbf{2.26}  & 0.77 & 0.87 & 0.82 \\
        Poisson skel. (w=0.05)  & 12.88 & 12.72 & 10.18  & 3.33 & 3.40 & 1.54 \\
        Poisson skel. (w=0.10)  & 15.02 & 15.41 & 10.58  & 3.67 & 3.89 & 1.89 \\
        Poisson skel. (w=0.20)  & 16.83 & 17.20 & 8.67  & 4.07 & 4.33 & 1.85 \\
        \hline
        CPMA  & 6.62 & 6.14 & 2.64  & 0.96 & 1.06 & 0.77 \\
        C-CPMA  & 6.19 & 5.77 & 3.30  & 0.98 & 1.05 & 0.86 \\
        \hline
    \end{tabular}
    \caption*{Rotation equivariance results on Animal2000. The table shows the average Hausdorff distance and Dubuisson-Jain dissimilarity for different rotations of each element in the dataset.}
\end{table}

    

Finally, we conducted the rotation sensitivity analysis on the 3D dataset. the results are summarized in Figure \ref{fig:rotation_results} (bottom row). The image shows the four 3D medial axis extraction methods we compared in our study for combinations of azimuthal and elevation angles. This figure shows how both the Hausdorff distance and the Dubuisson-Jain dissimilarity become higher when the rotation becomes more extreme, except in the case of C-CPMA. We believe this behavior is due to the connectivity enforcement mitigating the gaps in the medial axis, and reducing the metrics.

    

\subsection{Hyper-parameter Selection}
\label{subsec:hyper_parameter_selection}

We tested our medial axis pruning method on the Kimia216 dataset. Many medial axis pruning methods depend on hyper-parameters to accurately estimate the medial axis~\cite{Hesselink2008,Couprie2007,Giesen2009,Postolski2014}. These parameters usually have a physical meaning in the context of the object whose medial axis they seek to estimate. Often, the parameters are distances or angles formed between points inside the object. In other work, some authors create score functions like ours, intending to use its values as a filter parameter to remove points on spurious branches of the $\mathbf{MAT}$. In most cases, however,  such parameters are subject to factors like resolution or scale. Thus, we conducted another experiment to test the sensitivity of the CPMA to the pruning parameter $\tau$ at different scale factors of the input object. 

\begin{figure}[!ht]
    
    \centering
    \includegraphics[width=0.8\linewidth]{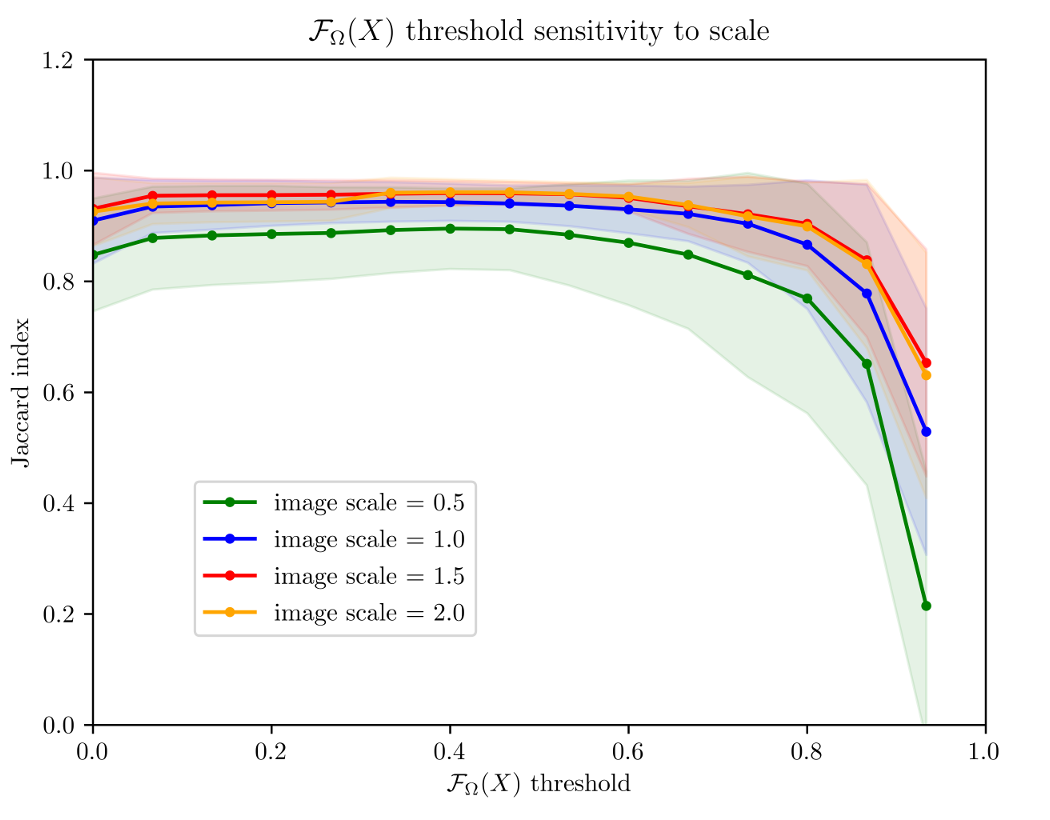}
    
    \caption{Sensitivity analysis of threshold $\tau$ to different scales of the input images. The graph shows the average Jaccard index of the reconstructed shape w.r.t the original object for the CPMA computed with different values of $\tau$. Higher values of the threshold lead to less spurious branches. We also show the standard deviation error bands.}
     \label{fig:scale_sensitivity_analysis}
\end{figure}

Figure \ref{fig:scale_sensitivity_analysis} shows the average of the Jaccard index as the reconstruction metric vs the values of $\tau$. We compared an object $\Omega$ against its reconstruction $\hat \Omega$ over all images in Kimia216 dataset. The figure shows how high values of $\tau$ deteriorate the reconstruction, while lower values do not prune enough spurious branches. From the figure, we can infer that values around $\tau=0.47$ offer a good trade-off between reconstruction and branch pruning. Moreover, around this value, the standard deviation reaches its minimum value suggesting optimal performance regardless of the object. Because the value of $\tau$ is stable for different scale factors, we conclude that scale does not affect the selection of the threshold.

%% file: sections/06_conclusions.tex
\label{sec:conclusions} 

We presented the CPMA, a new method for medial axis pruning with noise robustness and equivariance to isometric transformations. Our method leverages the discrete cosine transform to compute a score function that rates the importance of individual points and branches within the medial axis of a shape. 

Our pruning approach delivers competitive results compared to the state-of-the-art. The results of our experiments show that our method is robust to boundary noise. Additionally, it is also equivariant to isometric transformations, and it is capable of producing a stable and connected medial axis even in scenarios with significant perturbations of the contour. 

The CPMA can be efficiently computed in parallel because it depends on an aggregation of reconstructions of the original shape. Each reconstruction is independent of the others, which allows the parallelism. 

We believe our work leaves room for improvement. Thus we identified the following potential for future work.

All of the 3D objects we used come as 3D triangular meshes. To compute the CPMA, we discretize the meshes to fix a resolution of $150^3$ voxels. The discretization introduces two issues: 1) the objects lose small details in their structure, affecting the overall performance, and 2) the isometric equivariance decreases because rotated voxels will not usually align perfectly with non-rotated voxels. 

Our algorithm for connectivity enforcement relies on iterative computations of Dijkstra's algorithm for finding the minimum energy path between two pieces of the unconnected medial axis. Better strategies to compute the paths could increase the efficiency. 

%% file: main.bbl
\begin{thebibliography}{10}

\bibitem{Abbasi1999}
Sadegh Abbasi, Farzin Mokhtarian, and Josef Kittler.
\newblock Curvature scale space image in shape similarity retrieval.
\newblock {\em Multimedia Systems}, 7(6):467--476, November 1999.

\bibitem{ARCELLI1988361}
Carlo Arcelli and Gabriella~Sanniti di~Baja.
\newblock Finding local maxima in a pseudo-euclidian distance transform.
\newblock {\em Computer Vision, Graphics, and Image Processing}, 43(3):361 --
  367, 1988.

\bibitem{Arcelli2011}
Carlo Arcelli, Gabriella~Sanniti di~Baja, and Luca Serino.
\newblock {Distance-Driven Skeletonization in Voxel Images}.
\newblock {\em IEEE Transactions on Pattern Analysis and Machine Intelligence},
  33(4):709--720, apr 2011.

\bibitem{Aubert2014}
Gilles Aubert and Jean~Franois Aujol.
\newblock {Poisson skeleton revisited: A new mathematical perspective}.
\newblock {\em Journal of Mathematical Imaging and Vision}, 48(1):149--159,
  2014.

\bibitem{7071593}
G.~{Aufort}, R.~{Jennane}, R.~{Harba}, and C.~L. {Benhamou}.
\newblock A new 3d shape-dependant skeletonization method. application to
  porous media.
\newblock In {\em 2006 14th European Signal Processing Conference}, pages 1--5,
  2006.

\bibitem{August99}
J.~{August}, A.~{Tannembaum}, and S.~W. {Zucker}.
\newblock On the evolution of the skeleton.
\newblock In {\em Proceedings of the ICCV}, pages 315--322, 1999.

\bibitem{Bai2007ab}
X.~{Bai}, L.~J. {Latecki}, and W.~{Liu}.
\newblock Skeleton pruning by contour partitioning with discrete curve
  evolution.
\newblock {\em IEEE Transactions on Pattern Analysis and Machine Intelligence},
  29(3):449--462, 2007.

\bibitem{Bai2009}
X.~{Bai}, W.~{Liu}, and Z.~{Tu}.
\newblock Integrating contour and skeleton for shape classification.
\newblock In {\em 2009 IEEE 12th International Conference on Computer Vision
  Workshops, ICCV Workshops}, pages 360--367, 2009.

\bibitem{Belongie2002}
Serge Belongie, Jitendra Malik, and Jan Puzicha.
\newblock {Shape matching and object recognition using shape contexts}.
\newblock {\em IEEE Transactions on Pattern Analysis and Machine Intelligence},
  24(4):509--522, 2002.

\bibitem{5395557}
A.~{Beristain} and M.~{Grana}.
\newblock Pruning algorithm for voronoi skeletons.
\newblock {\em Electronics Letters}, 46(1):39--41, January 2010.

\bibitem{Blum1967}
Harry Blum.
\newblock {A} {T}ransformation for {E}xtracting {N}ew {D}escriptors of {S}hape.
\newblock In Weiant Wathen-Dunn, editor, {\em Models for the Perception of
  Speech and Visual Form}, pages 362--380. MIT Press, Cambridge, 1967.

\bibitem{Chaudhry2013}
Rizwan Chaudhry, Ferda Ofli, Gregorij Kurillo, Ruzena Bajcsy, and Rene Vidal.
\newblock Bio-inspired dynamic 3d discriminative skeletal features for human
  action recognition.
\newblock In {\em 2013 {IEEE} Conference on Computer Vision and Pattern
  Recognition Workshops}, pages 471--478. {IEEE}, June 2013.

\bibitem{Chaussard2011}
John Chaussard, Michel Couprie, and Hugues Talbot.
\newblock {Robust skeletonization using the discrete $\lambda$-medial axis}.
\newblock {\em Pattern Recognition Letters}, 32(9):1384--1394, jul 2011.

\bibitem{Couprie2007}
Michel Couprie, David Coeurjolly, and Rita Zrour.
\newblock {Discrete bisector function and Euclidean skeleton in 2D and 3D}.
\newblock {\em Image and Vision Computing}, 25(10):1543--1556, 2007.

\bibitem{Dubuisson2002}
M.-P. Dubuisson and A.K. Jain.
\newblock {A modified Hausdorff distance for object matching}.
\newblock In {\em Proceedings of 12th International Conference on Pattern
  Recognition}, volume~1, pages 566--568. IEEE Comput. Soc. Press, 2002.

\bibitem{Dlotko14}
P.~{Dłotko} and R.~{Specogna}.
\newblock Topology preserving thinning of cell complexes.
\newblock {\em IEEE Transactions on Image Processing}, 23(10):4486--4495, 2014.

\bibitem{Esteves2018a}
Carlos Esteves, Christine Allen-Blanchette, Ameesh Makadia, and Kostas
  Daniilidis.
\newblock Learning so(3) equivariant representations with spherical cnns.
\newblock In Vittorio Ferrari, Martial Hebert, Cristian Sminchisescu, and Yair
  Weiss, editors, {\em Computer Vision -- ECCV 2018}, pages 54--70, Cham, 2018.
  Springer International Publishing.

\bibitem{Black2016}
Oren Freifeld and Michael~J Black.
\newblock {Lie Bodies: A Manifold Representation of 3D Human Shape}.
\newblock In Bastian Leibe, Jiri Matas, Nicu Sebe, and Max Welling, editors,
  {\em European Conference on Computer Vision}, number October 2012 in Lecture
  Notes in Computer Science, pages 1--14. Springer International Publishing,
  Cham, 2012.

\bibitem{Gao2018}
Fengyi Gao, Guangshun Wei, Shiqing Xin, Shanshan Gao, and Yuanfeng Zhou.
\newblock {2D skeleton extraction based on heat equation}.
\newblock {\em Computers and Graphics (Pergamon)}, 74:99--108, 2018.

\bibitem{Giesen2009}
Joachim Giesen, Balint Miklos, Mark Pauly, and Camille Wormser.
\newblock {The scale axis transform}.
\newblock In {\em Proceedings of the 25th annual symposium on Computational
  geometry - SCG '09}, page 106, New York, New York, USA, 2009. ACM Press.

\bibitem{Gorelick2006}
Lena Gorelick, Meirav Galun, and Eitan Sharon.
\newblock {Shape Representation and Classification Using the Poisson Equation}.
\newblock {\em IEEE Transactions on Pattern Analysis and Machine Intelligence},
  28(12):1991--2005, 2006.

\bibitem{Hesselink2008}
Wim~H. Hesselink and Jos~B.T.M. Roerdink.
\newblock {Euclidean skeletons of digital image and volume data in linear time
  by the integer medial axis transform}.
\newblock {\em IEEE Transactions on Pattern Analysis and Machine Intelligence},
  30(12):2204--2217, 2008.

\bibitem{Li2018}
Haisheng Li, Li~Sun, Xiaoqun Wu, and Qiang Cai.
\newblock Scale-invariant wave kernel signature for non-rigid 3d shape
  retrieval.
\newblock In {\em 2018 {IEEE} International Conference on Big Data and Smart
  Computing ({BigComp})}, pages 448--454. {IEEE}, January 2018.

\bibitem{Li2019ActionalStructuralGC}
Maosen Li, Siheng Chen, Xu~Chen, Ya~Zhang, Yanfeng Wang, and Qi~Tian.
\newblock Actional-structural graph convolutional networks for skeleton-based
  action recognition.
\newblock {\em ArXiv}, abs/1904.12659, 2019.

\bibitem{Li2017abc}
Ronghao Li, Guochao Bu, and Pei Wang.
\newblock An automatic tree skeleton extracting method based on point cloud of
  terrestrial laser scanner.
\newblock {\em International Journal of Optics}, 2017:1--11, 2017.

\bibitem{5396343}
C.~{Lohou} and J.~{Dehos}.
\newblock Automatic correction of ma and sonka's thinning algorithm using
  p-simple points.
\newblock {\em IEEE Transactions on Pattern Analysis and Machine Intelligence},
  32(6):1148--1152, 2010.

\bibitem{LOHOU2004171}
Christophe Lohou and Gilles Bertrand.
\newblock A 3d 12-subiteration thinning algorithm based on p-simple points.
\newblock {\em Discrete Applied Mathematics}, 139(1):171 -- 195, 2004.
\newblock The 2001 International Workshop on Combinatorial Image Analysis.

\bibitem{Marie2016}
Romain Marie, Ouiddad Labbani-Igbida, and El~Mustapha Mouaddib.
\newblock {The Delta Medial Axis: A fast and robust algorithm for filtered
  skeleton extraction}.
\newblock {\em Pattern Recognition}, 56:26--39, 2016.

\bibitem{Maturana2015}
D.~{Maturana} and S.~{Scherer}.
\newblock Voxnet: A 3d convolutional neural network for real-time object
  recognition.
\newblock In {\em 2015 IEEE/RSJ International Conference on Intelligent Robots
  and Systems (IROS)}, pages 922--928, Sep. 2015.

\bibitem{Miklos:2010:DSA:1778765.1778838}
Balint Miklos, Joachim Giesen, and Mark Pauly.
\newblock Discrete scale axis representations for 3d geometry.
\newblock {\em ACM Trans. Graph.}, 29(4):101:1--101:10, July 2010.

\bibitem{Mokhtarian1992789}
F.~Mokhtarian and A.K. Mackworth.
\newblock A theory of multiscale, curvature-based shape representation for
  planar curves.
\newblock {\em IEEE Transactions on Pattern Analysis and Machine Intelligence},
  14(8):789--805, 1992.
\newblock cited By 722.

\bibitem{Nemeth2011}
G{\'{a}}bor N{\'{e}}meth, P{\'{e}}ter Kardos, and K{\'{a}}lm{\'{a}}n
  Pal{\'{a}}gyi.
\newblock {Thinning combined with iteration-by-iteration smoothing for 3D
  binary images}.
\newblock {\em Graphical Models}, 73(6):335--345, 2011.

\bibitem{Ogniewicz1992}
R.~{Ogniewicz} and M.~{Ilg}.
\newblock Voronoi skeletons: theory and applications.
\newblock In {\em Proceedings 1992 IEEE Computer Society Conference on Computer
  Vision and Pattern Recognition}, pages 63--69, June 1992.

\bibitem{Peters2016}
Ravi Peters and Hugo Ledoux.
\newblock Robust approximation of the medial axis transform of {LiDAR} point
  clouds as a tool for visualisation.
\newblock {\em Computers {\&} Geosciences}, 90:123--133, May 2016.

\bibitem{Postolski2014}
Micha{\l} Postolski, Michel Couprie, and Marcin Janaszewski.
\newblock {Scale filtered Euclidean medial axis and its hierarchy}.
\newblock {\em Computer Vision and Image Understanding}, 129:89--102, 2014.

\bibitem{Saha2017}
Gunilla~Borgefors Punam K.~Saha and Gabriella~Sanniti de~Baja~(Eds.).
\newblock {\em Skeletonization. Theory, Methods and Applications}.
\newblock Elsevier Science, 1st edition edition, 2017.

\bibitem{6005551}
T.~{Qiu}, Y.~{Yan}, and G.~{Lu}.
\newblock A medial axis extraction algorithm for the processing of combustion
  flame images.
\newblock In {\em 2011 Sixth International Conference on Image and Graphics},
  pages 182--186, Aug 2011.

\bibitem{Rumpf2002}
Martin Rumpf and Tobias Preusser.
\newblock A level set method for anisotropic geometric diffusion in 3d image
  processing.
\newblock {\em {SIAM} Journal on Applied Mathematics}, 62(5):1772--1793,
  January 2002.

\bibitem{Safar2003}
Maytham~H. Safar and Cyrus Shahabi.
\newblock {\em Shape Analysis and Retrieval of Multimedia Objects}.
\newblock Springer {US}, 2003.

\bibitem{Saha2016}
Punam~K. Saha, Gunilla Borgefors, and Gabriella {Sanniti di Baja}.
\newblock {A survey on skeletonization algorithms and their applications}.
\newblock {\em Pattern Recognition Letters}, 76:3--12, 2016.

\bibitem{Sato2000}
M.~{Sato}, I.~{Bitter}, M.~A. {Bender}, A.~E. {Kaufman}, and M.~{Nakajima}.
\newblock Teasar: tree-structure extraction algorithm for accurate and robust
  skeletons.
\newblock In {\em Proceedings the Eighth Pacific Conference on Computer
  Graphics and Applications}, pages 281--449, Oct 2000.

\bibitem{Sebastian2004}
Thomas~B. Sebastian, Philip~N. Klein, and Benjamin~B. Kimia.
\newblock {Recognition of shapes by editing shock graphs}.
\newblock {\em IEEE TRANSACTIONS ON PATTERN ANALYSIS AND MACHINE INTELLIGENCE},
  00(528):755--762, 2004.

\bibitem{SHAKED1998156}
Doron Shaked and Alfred~M. Bruckstein.
\newblock Pruning medial axes.
\newblock {\em Computer Vision and Image Understanding}, 69(2):156 -- 169,
  1998.

\bibitem{Shen2011AB}
Wei Shen, Xiang Bai, Rong Hu, Hongyuan Wang, and Longin Jan~Latecki.
\newblock Skeleton growing and pruning with bending potential ratio.
\newblock {\em Pattern Recogn.}, 44(2):196–209, February 2011.

\bibitem{Siddiqi1999}
Kaleem Siddiqi, Ali Shokoufandeh, Sven~J. Dickinson, and Steven~W. Zucker.
\newblock {Shock graphs and shape matching}.
\newblock {\em International Journal of Computer Vision}, 35(1):13--32, 1999.

\bibitem{Sobiecki2014}
Andr{\'{e}} Sobiecki, Andrei Jalba, and Alexandru Telea.
\newblock {Comparison of curve and surface skeletonization methods for voxel
  shapes}.
\newblock {\em Pattern Recognition Letters}, 47:147--156, oct 2014.

\bibitem{Sobiecki2013}
Andr{\'{e}} Sobiecki, Haluk~C. Yasan, Andrei~C. Jalba, and Alexandru~C. Telea.
\newblock {Qualitative Comparison of Contraction-Based Curve Skeletonization
  Methods}.
\newblock In {\em Lecture Notes in Computer Science (including subseries
  Lecture Notes in Artificial Intelligence and Lecture Notes in
  Bioinformatics)}, volume 7883 LNCS, pages 425--439. Springer, 2013.

\bibitem{Sun2009}
Jian Sun, Maks Ovsjanikov, and Leonidas Guibas.
\newblock {A Concise and Provably Informative Multi-Scale Signature Based on
  Heat Diffusion}.
\newblock {\em Computer Graphics Forum}, 28(5):1383--1392, jul 2009.

\bibitem{Tal2014}
Ayellet Tal.
\newblock 3d shape analysis for archaeology.
\newblock In {\em 3D Research Challenges in Cultural Heritage}, pages 50--63.
  Springer Berlin Heidelberg, 2014.

\bibitem{Toshev2012}
Alexander Toshev, Ben Taskar, and Kostas Daniilidis.
\newblock Shape-based object detection via boundary structure segmentation.
\newblock {\em International Journal of Computer Vision}, 99(2):123--146, March
  2012.

\bibitem{Maaten2006}
Laurens J.~P. van~der Maaten, Paul~J. Boon, Guus Lange, Hans Paijmans, and
  Eric~O. Postma.
\newblock Computer vision and machine learning for archaeology.
\newblock In {\em Proceedings of the Computer Applications in Archaeology, CAA
  2006}, page in press. Dr. H. Kamermans, Faculty of Archeology, Leiden
  University, 2006.

\bibitem{Viswanathan2013}
G.~K. {Viswanathan}, A.~{Murugesan}, and K.~{Nallaperumal}.
\newblock A parallel thinning algorithm for contour extraction and medial axis
  transform.
\newblock In {\em 2013 IEEE International Conference ON Emerging Trends in
  Computing, Communication and Nanotechnology (ICECCN)}, pages 606--610, March
  2013.

\bibitem{younes2019}
L.~Younes.
\newblock {\em Shapes and Diffeomorphisms}.
\newblock Applied Mathematical Sciences. Springer Berlin Heidelberg, 2 edition,
  2019.

\bibitem{Zhang2004ab}
Dengsheng Zhang and Guojun Lu.
\newblock Review of shape representation and description techniques.
\newblock {\em Pattern Recognition}, 37(1):1 -- 19, 2004.

\bibitem{Zhang1984}
T.~Y. Zhang and C.~Y. Suen.
\newblock A fast parallel algorithm for thinning digital patterns.
\newblock {\em Commun. ACM}, 27(3):236--239, March 1984.

\end{thebibliography}
